\documentclass[a4paper,fleqn]{cas-dc}

\usepackage{natbib}
\usepackage{caption}
\usepackage{bbding} 
\usepackage{amsmath}
\usepackage{hyperref}

\usepackage{graphicx}
\usepackage{bbding}
\usepackage{amsmath}
\usepackage{amsfonts}
\usepackage{enumerate}
\usepackage{booktabs}
\usepackage{algorithm}
\usepackage{algorithmic}
\usepackage{colortbl}
\usepackage{bbm}
\usepackage{multirow}
\usepackage{array}
\usepackage{pifont}

\newcommand{\etal}{\textit{et al.}}


\def\tsc#1{\csdef{#1}{\textsc{\lowercase{#1}}\xspace}}
\tsc{WGM}
\tsc{QE}

\begin{document}
\let\WriteBookmarks\relax
\def\floatpagepagefraction{1}
\def\textpagefraction{.001}

\shorttitle{}    
\shortauthors{Xu et al.}  

\title[mode = title]{Dual Selective Fusion Transformer Network for Hyperspectral Image Classification}



\author[1,2]{Yichu Xu}[style=chinese]
\ead{xuyichu@whu.edu.cn}

\address[1]{ Institute of Artificial Intelligence, School of Computer Science, Wuhan University, Wuhan,  430072, PR China}
\address[2]{  Hubei Luojia Laboratory, Wuhan 430079, PR China}
\address[3]{ Aerospace Information Research Institute, Henan Academy of Sciences, Zhengzhou 450046, PR China}

\author[1,2]{Di Wang}[style=chinese]
\ead{d\_wang@whu.edu.cn}


\author[1,2]{Lefei Zhang}[style=chinese]
\cormark[1]
\credit{Supervision.}
\ead{zhanglefei@whu.edu.cn}

\author[3]{Liangpei Zhang}[style=chinese]
\ead{zlp62@whu.edu.cn}

\cortext[1]{Corresponding author.}

\begin{abstract}
 Transformer has achieved satisfactory results in the field of hyperspectral image (HSI) classification. However, existing Transformer models face two key challenges when dealing with HSI scenes characterized by diverse land cover types and rich spectral information: (1) A fixed receptive field overlooks the effective contextual scales required by various HSI objects; (2) invalid self-attention features in context fusion affect model performance. To address these limitations, we propose a novel Dual Selective Fusion Transformer Network (DSFormer) for HSI classification. DSFormer achieves joint spatial and spectral contextual modeling by flexibly selecting and fusing features across different receptive fields, effectively reducing unnecessary information interference by focusing on the most relevant spatial-spectral tokens. Specifically, we design a Kernel Selective Fusion Transformer Block (KSFTB) to learn an optimal receptive field by adaptively fusing spatial and spectral features across different scales, enhancing the model’s ability to accurately identify diverse HSI objects. Additionally, we introduce a Token Selective Fusion Transformer Block (TSFTB), which strategically selects and combines essential tokens during the spatial-spectral self-attention fusion process to capture the most crucial contexts. Extensive experiments conducted on four benchmark HSI datasets demonstrate that the proposed DSFormer significantly improves land cover classification accuracy, outperforming existing state-of-the-art methods. Specifically, DSFormer achieves overall accuracies of 96.59\%, 97.66\%, 95.17\%, and 94.59\% in the Pavia University, Houston, Indian Pines, and Whu-HongHu datasets, respectively, reflecting improvements of 3.19\%, 1.14\%, 0.91\%, and 2.80\% over the previous model. The code will be available online at https://github.com/YichuXu/DSFormer.
\end{abstract}
\begin{keywords}
\sep Hyperspectral image classification \sep Transformer \sep Receptive field \sep Self-attention \sep Spatial-spectral joint
\end{keywords}



\maketitle
\begin{sloppypar}

\section{Introduction}\label{introduction}
With the continuous advancements in hyperspectral imaging technology, hyperspectral images (HSIs) now offer increasingly rich spatial-spectral information, enabling precise Earth Observation \citep{YangNN24,zhang2022rss,paoletti2019deep,xu2023ai,hypersigma}. Leveraging their distinctive spectral characteristics, HSIs are capable of performing a wide range of tasks, including classification \citep{ShiNN23,li2019Deep,Gao24FAI,roy2023}, target detection \citep{ZhuNN23,Xu2022HAD,XieNN21}, change detection \citep{Change2Cap,BCG-HCD,chen2024changemamba}, and image quality enhancement \citep{xiao2024fmsr,WangNN22,msdformer}. Notably, HSI classification has become a fundamental task in the remote sensing community due to its extensive applications in areas such as disaster monitoring \citep{Hong_hsi_grsm}, precision agriculture \citep{agriculture}, and urban planning \citep{hsi_urban_plan}. The primary goal of HSI classification is to assign specific class labels to each pixel within the HSI. 

HSI classification increasingly relies on deep learning techniques, which offer significant advantages over traditional machine learning methods due to their powerful feature extraction capabilities \citep{2D-CNN,mou_cnn}. In particularly, convolutional neural networks (CNNs) enable the extraction of features in a hierarchical manner \citep{HybridSN,rpnet}. However, their local convolutional structure limits their ability to capture long-range dependencies. To address this, recent studies \citep{wang2018non} introduced self-attention mechanisms to capture non-local contextual information, though at a high computational cost due to extensive similarity calculations.

Recently, the success of self-attention in natural language processing \citep{selfattention} has inspired its application in computer vision. Dosovitskiy~\etal~\citep{vit2021} proposed the Vision Transformer (ViT), which segments images into tokens and applies self-attention to capture long-range dependencies, achieving notable improvements in image classification. Transformer-based approaches \citep{HMSST,SSTN,SpectralFormer,GSCViT} have become a mainstream trend in HSI classification research, with numerous studies introducing novel methods to advance HSI classification technology.

\begin{figure}[t]
\centering
\includegraphics[width=\linewidth]{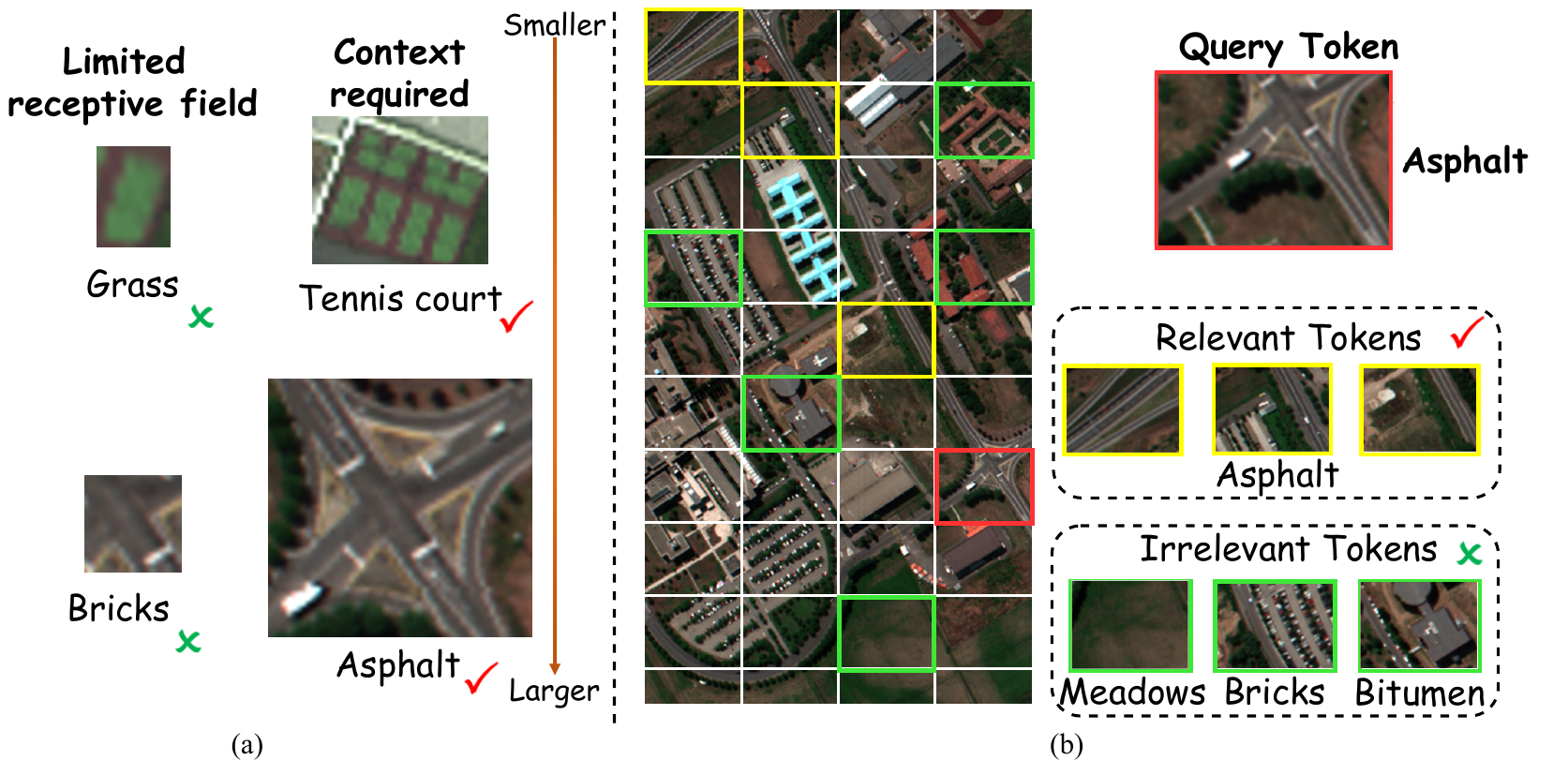}
\caption{ The challenges existing in current HSI classification methods. (a) Fixed and limited receptive fields in HSI classification can potentially lead to misclassifications, such as confusing asphalt to bricks. (b) 
Irrelevant tokens may introduce noise which may potentially impair classification accuracy.}
\label{fig:Intro}
\end{figure}

However, most transformer structures typically utilize a fixed receptive field for feature extraction, overlooking the fact that different types of land cover require varying contextual information \citep{li2019selective,li2023large}. As shown in Fig.~\ref{fig:Intro} (a), HSI with a fixed receptive field may exhibit misclassification issues. Therefore, we argue that different objects in HSIs should be recognized with correspondingly required receptive fields. After determining the scope of visual perception, the next step is to perform effective context capturing.
To achieve this, most transformer-based HSI classification models usually employ vanilla Multi-Head Self-Attention (MHSA) to build global features \citep{GAHT} by enabling interactions between one token and all the other tokens. However, since not all tokens are informative, the original MHSA inevitably introduces unexpected noise during context feature extraction. As shown in Fig.~\ref{fig:Intro} (b), the green boxes represent unnecessary tokens relative to the red query marker for asphalt, such as meadows and bitumen. 
Additionally, in capturing contextual information, existing transformer-based HSI classification models predominantly focus on either spatial or spectral information alone \citep{SpectralFormer}, neglecting the integrated modeling of spatial-spectral contexts within HSI data.

To address these issues, this paper proposes a novel Dual Selective Fusion Transformer Network (DSFormer) for HSI classification, which selects and fuses the most valuable and relevant spatial-spectral information within a reasonable receptive field. 
To accommodate this, DSFormer incorporates a Kernel Selective Fusion Transformer Block (KSFTB), which dynamically selects and integrates multiple convolutional kernels of appropriate sizes, optimizing the receptive fields to obtain effective spatial-spectral features. Furthermore, rather than utilizing all tokens for dense attention calculation, DSFormer introduces a Token Selective Fusion Transformer Block (TSFTB). This block selects the most relevant tokens of the highest attention values for fusion through a spectral grouping strategy that can directly utilize the 3D HSI characteristic. These selective mechanisms allow DSFormer to produce the most significant and appropriate spatial-spectral information within an optimal receptive field. The main contributions of this study are summarized as follows:

\begin{enumerate}[(1)]
\item This paper proposes a novel Dual Selective Fusion Transformer Network (DSFormer) for HSI classification. It adaptively selects and fuses features from diverse receptive fields to achieve joint spatial-spectral context modeling, while reducing unnecessary information interference by focusing on the most relevant spatial-spectral tokens. This approach enables the precise classification of HSI ground objects.

\item To adaptively select and integrate the appropriate contextual information, we design a Kernel Selective Fusion Transformer Block (KSFTB). This mechanism ensures that features are acquired within an optimal receptive field from both spatial and spectral perspectives, thereby enhancing the model's capacity to effectively perceive and distinguish various HSI objects.

\item  We further develop a Token Selective Fusion Transformer Block (TSFTB) that strategically selects the most relevant tokens in the context fusion process, effectively integrating essential information while reducing the impact of irrelevant objects, thereby improving the distinguishability of the generated spatial-spectral features for HSI visual interpretation.

\item {Extensive experiments are conducted on four benchmark HSI datasets: Pavia University, Houston, Indian Pines, and Whu-HongHu.} The results indicate that the proposed DSFormer, which utilizes appropriate and valuable contextual information, performs better than other state-of-the-art methods.
\end{enumerate}

\par The remaining section of this paper is organized as follows. Section \ref{Section2} reviews the related work of HSI classification. In Section \ref{Section3} provides a comprehensive description of the proposed DSFormer model. Section \ref{Section4} presents and thoroughly analyzes the experimental results. Finally, Section \ref{Section5} concludes the paper.

\section{Related Works} \label{Section2}

\subsection{CNN-based Methods}
With the rapid progress in deep learning, CNNs have shown remarkable potential and significant practical value across a wide range of computer vision tasks, particularly in the field of HSI classification. Owing to their powerful feature representation capabilities, CNNs \citep{zhao_cnn,Zhu2017grsm,chen_cnn,ssfcn} have attracted significant attention. A 1D-CNN \citep{hu2015deep} was proposed to achieve efficient local spectral feature extraction through multiple convolutional layers. Further, 2D-CNN was introduced in \citep{2D-CNN} that extracts spatial-spectral features and uses a balanced local discriminant embedding. Later, 3D-CNNs \citep{3d-cnn,li2017spectral} were developed to directly extract joint spatial-spectral features from the HSI cube. Then, some multiscale convolutional networks \citep{ssun,MS-CNN,wang_Assmn,yang2021enhanced} were proposed to extract deep multiscale features.

\subsection{Transformer-based Methods}
However, the above CNN-based networks are constrained by their inherent local perception, which limits the amount of information they can capture, and thus hinders accuracy. Subsequently, HSI classification approaches with attention mechanism \citep{SSAN,AA-CNN,RSSAN,Fcnet} emerged and achieved significant classification results, particularly self-attention mechanisms \citep{selfattention,SACNet,Xu2024s3anet}, due to their ability to improve long-range perception.

\begin{figure*}[!htb]
\centering
\includegraphics[width=\linewidth]{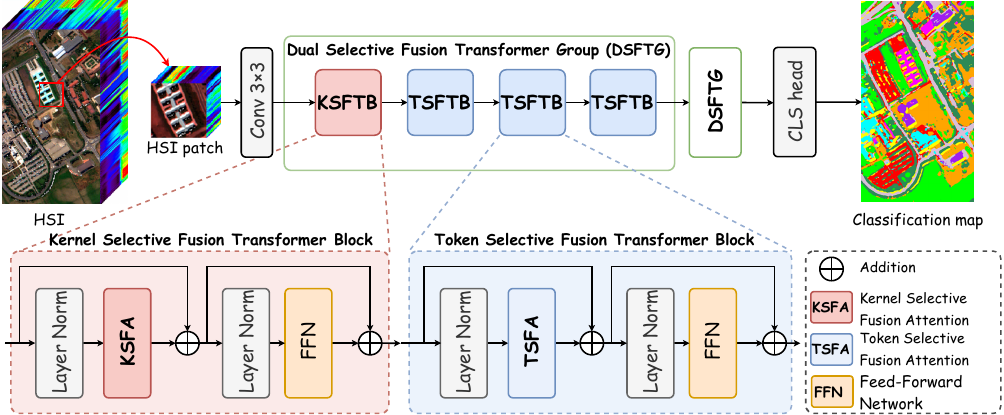}
\caption{ An illustration of the proposed DSFormer. The Dual Selective Fusion Transformer Group (DSFTG) is composed of a Kernel Selective Fusion Transformer Block (KSFTB) and three consecutive Token Selective Fusion Transformer Blocks (TSFTBs). Followed by two DSFTGs, we adopt the fully connected layer as the classification head to obtain the classification results.}
\label{fig:Framework}
\end{figure*}

Transformer-based networks \citep{chen2023learning,ttst,SSFTT,CACFT,MASSFormer} have been widely adopted for their use of MHSA to achieve superior context modeling. These approaches have notably outperformed traditional CNNs, which are limited by their receptive fields. SpectralFormer was proposed in \citep{SpectralFormer}, a network that enhances HSI classification by capturing spectral sequence information between adjacent HSI bands using transformers and cross-layer skip connections. Further, Sun~\etal~\citep{SSFTT} proposed a spectral–spatial feature tokenization transformer (SSFTT), where shallow features are extracted by convolutions, semantic features are processed by a Gaussian weighted tokenization module, and a single-layer transformer is used for feature representation. Then, LESSFormer was introduced \citep{Lessformer}, which employs an HSI2Token module to extract adaptive spatial-spectral tokens and uses a local transformer to enhance the local spectral representation. Inspired by the effectiveness of spectral partitioning, Mei introduced a grouped pixel embedding strategy and hierarchical structure to extract discriminative multiscale spatial-spectral features \citep{GAHT}.  Yang~\etal~\citep{yang2023center} introduced a transformer-based center-to-surround interactive learning (CSIL) framework for extracting multi-scale features for HSI classification tasks. Roy~\etal~\citep{morphformer} incorporated morphological characteristics into their proposed MorphFormer by combining spectral and spatial morphological convolution operations with attention mechanisms to effectively integrate structural and morphological information. After that, Zhao~\etal~\citep{GSCViT} introduced a lightweight variant of the ViT network, which combines group-separable convolution and MHSA modules within feature extraction blocks to efficiently capture local and global spatial features with reduced parameter complexity.

Compared with the aforementioned Transformer-based methods, we contend that the proposed DSFormer is distinct. First, KSFTB is specifically designed to adaptively choose and fuse relevant context scopes. This mechanism ensures that feature extraction takes place within an optimal receptive field, encompassing both spatial and spectral dimensions. We further develop TSFTB, which aims to combine the most relevant tokens to capture valuable information. This mechanism effectively integrates essential information while mitigating the influence of redundant and irrelevant ingredients, thereby enhancing the representation capability of spatial-spectral features for HSI classification.

\section{Methodology}\label{Section3}
In this section, we present the implementation details of our DSFormer. DSFormer consists of convolutional layers, dual selective fusion transformer groups (DSFTGs), and a classification head. We begin with an overview of DSFormer, followed by a detailed explanation of two components KSFTB, and TSFTB.

\begin{figure}[t]
\centering
\includegraphics[width=\linewidth]{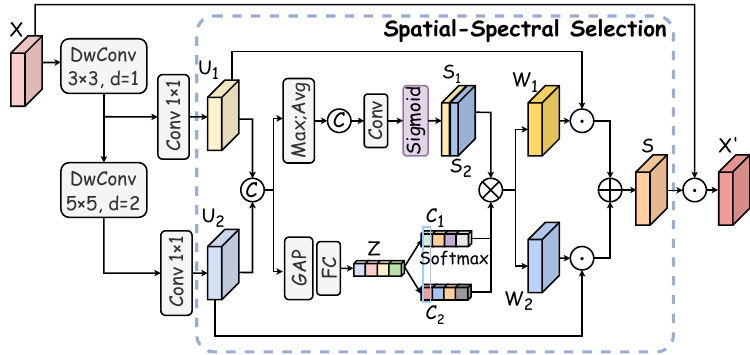}
\caption{The proposed Kernel Selective Fusion Attention (KSFA). Here, DwConv represents Depth-wise convolution, Avg and Max represent channel-wise average pooling and max pooling, respectively, and GAP means spatial global average pooling.}
\label{fig:KSFA}
\end{figure}

\subsection{Overview of DSFormer}

As shown in Fig.~\ref{fig:Framework}, the HSI is first subjected to PCA for dimensionality reduction, followed by the extraction of overlapping patch cubes, \(\{ x_i \in \mathbb{R}^{c \times w \times w} \}_{i=1}^{H \times W}\), where each patch is classified based on the central pixel. Here, \(H\) and \(W\) denote the height and width, $w$ represents the patch size, and \(c\) is the reduced spectral dimension. Taking the pixel-level patch cube \(x \in \mathbb{R}^{c \times w \times w}\) as input, we first employ a 3×3 convolutional layer to capture shallow features, denoted as $\mathbf{F}_1$. Next, we perform deep feature generation on $\mathbf{F}_1$ using two consecutive DSFTGs. This process can be formulated as: \(\mathbf{F}_2 = \text{DSFTG}_2(\text{DSFTG}_1(\mathbf{F}_1))\). Finally, a layer normalization followed by a fully connected layer in the classification head is applied to obtain the final classification results.

\subsection{Kernel Selective Fusion Transformer Block}

As analyzed in the Introduction, different land cover classes in HSI require distinct contextual scales. Existing methods are often limited by the fixed and limited size of the receptive field of the network or cannot dynamically adapt the receptive field to suit specific types of land cover \cite{SSFTT,morphformer}. To address this issue, we designed the KSFTB which learns an optimal receptive field through selecting and integrating multiscale features. KSFTB initially employs dilated depthwise convolutions to construct a large receptive field to obtain sufficient range of feature perception. Then, it incorporates a spatial-spectral selection mechanism to adaptively fusion the appropriate receptive field size required for different types of land cover. Specifically, KSFTB primarily consists of a Kernel Selective Fusion Attention (KSFA) module and a Feed-Forward Network (FFN).

The structure of KSFA is shown in Fig.~\ref{fig:KSFA}, two decoupled depth-wise convolutions (DwConv) are employed for input $\mathbf{X} \in \mathbb{R}^{c \times w \times w}$. The features generated with decomposed kernels are then processed by 1×1 convolution layers \(\mathcal{F}_{1 \times 1}(\cdot)\) to integrate channel information and unify dimensions, obtaining contextual features with different receptive fields:
\begin{equation}
    \mathbf{{U}_1} = \mathcal{F}^1_{1 \times 1} (\mathcal{F}_{3 \times 3}^{dwc}(\mathbf{X})),
    \mathbf{{U}_2} = \mathcal{F}^2_{1 \times 1} (\mathcal{F}_{5 \times 5}^{dwc}(\mathcal{F}_{3 \times 3}^{dwc}(\mathbf{X}))),
    \label{eqn:U1}
\end{equation}
where $\mathcal{F}_{3 \times 3}^{dwc}$ and $\mathcal{F}_{5 \times 5}^{dwc}$ denote a 3×3 DwConv and a 5×5 DwConv with a dilation of 2, respectively. 

Given the varying scales of contextual information required by different types of land cover, we employ a spatial-spectral selection mechanism to enhance the dynamic adaptability of the network. Specifically, we first concatenate the features $\mathbf{{U}_1},\mathbf{{U}_2} \in \mathbb{R}^{\frac{c}{2} \times w \times w}$ extracted by convolutional kernels with diverse receptive fields:
\begin{equation}
    \mathbf{U} = \left[\mathbf{{U}_1}; \mathbf{{U}_2} \right] \in \mathbb{R}^{c \times w \times w},
    \label{eqn:U}
\end{equation}
subsequently, channel-wise average pooling $\mathcal{P}_{\text {avg }}(\cdot)$ and max pooling $\mathcal{P}_{\text {max}}(\cdot)$ are applied to squeeze $\mathbf{U}$ to capture spatial descriptors. To enable information exchange among different spatial descriptors, we first concatenate the aforementioned pooling features and then using a simple 1×1 convolutional layer $\mathcal{F}^{2 \rightarrow O}_{1 \times 1}$ to transform the two pooling features into $O$ spatial attention maps, we set $O$ to 2 in our structure. These process can be written as below:
\begin{equation}
    \mathcal{\Bar{\mathbf{S}}}=\mathcal{F}^{2 \rightarrow O}_{1 \times 1}(\left[\mathcal{P}_{\text {avg}}(\mathbf{U}); \mathcal{P}_{\text {max}}(\mathbf{U})\right]).
    \label{eqn:SA}
\end{equation}

Following this, the sigmoid activation function $\sigma(\cdot)$ is applied to each spatial attention feature map $\mathcal{\Bar{\mathbf{S}}} \in \mathbb{R}^{2 \times w \times w} $, generating separate spatial selection masks corresponding to the different convolutional kernels:
\begin{equation}
    \mathbf{S}_i = \sigma(\mathcal{\Bar{\mathbf{S}}}_i).
    \label{eqn:S}
\end{equation}

Considering the substantial differences in contextual spectral information required by various types of complex land cover in HSIs, we have devised a spectral selection mechanism to meet this demand. First, we utilize spatial global average pooling $\mathcal{P}_{\text {GAP }}(\cdot)$ to compress the feature $\mathbf{U}$ into spectral feature descriptors. Next, a fully connected layer $\mathcal{F}_{\text {FC }}(\cdot)$ is employed to produce more compact spectral attention features $\boldsymbol{Z} \in \mathbb{R}^{c \times 1} $. These operations can be mathematically expressed as follows:
\begin{equation}
    \boldsymbol{Z} = \mathcal{F}_{\text {FC }}(\mathcal{P}_{\text {GAP }}(\mathbf{U})).
    \label{eqn:Z}
\end{equation}

Subsequently, spectral-wise attention is generated through a softmax operation to guide the selection process, resulting in spectral selection masks, which can be obtained as follows:
\begin{equation}
    C_{1}^ {[j]}=\frac{e^{{M_{j}} {Z_{j}}}}{e^{{M_{j}} {Z_{j}}}+e^{{N_{j}} {Z_{j}}}}, C_{2}^ {[j]}=\frac{e^{{N_{j}} {Z_{j}}}}{e^{{M_{j}} {Z_{j}}}+e^{{N_{j}} {Z_{j}}}},
    \label{eqn:C}
\end{equation}
where $C_i^{[j]}$ represents $j$-th attention value in $\boldsymbol{C_i} \in \mathbb{R}^{c \times 1},i=1,2$, $M_{j}$ refers the $j$-th row of $\boldsymbol{M}$, likewise $C_{2j}$ and $N_{j}$, and $\boldsymbol{M}, \boldsymbol{N} \in \mathbb{R}^{c \times 1}$ are two initially defined learnable vectors.

Furthermore, the spatial-spectral selection weights are obtained by matrix multiplication with the corresponding spatial-spectral selection masks:
\begin{equation}
    \mathbf{W}_1 = \boldsymbol{C}_1  \times \mathbf{S}_1  , \mathbf{W}_2 = \boldsymbol{C}_2 \times \mathbf{S}_2  .
    \label{eqn:W}
\end{equation}

The enhanced feature maps derived from the decomposed kernel sequence are weighted using their respective spatial-spectral selection weights $\mathbf{W}_{i} \in \mathbb{R}^{c \times w \times w }$. These weighted maps are then combined via the convolutional layer $\mathcal{F}_{1 \times 1}(\cdot), i=1,2$, the attention feature $\mathbf{S}$ can be expressed as:
\begin{equation}
    \mathbf{S} = \mathcal{F}_{1 \times 1}(\sum_{i=1}^{O} \left ( \mathbf{W}_i \cdot \mathbf{U}_i \right ) ).
    \label{eqn:SS}
\end{equation}

The KSFA module generates its final output by performing an element-wise multiplication between the input features $\mathbf{X} $ and $\mathbf{S} $:
\begin{equation}
    \mathbf{\mathbf{{X}^{'}}} =\mathbf{X}\cdot \mathbf{S}.
    \label{eqn:X}
\end{equation}

The FFN module comprises a fully connected layer, a depthwise convolution, a GELU activation function, and a second fully connected layer.

\subsection{Token Selective Fusion Transformer Block}

\begin{figure}[t]
\centering
\includegraphics[width=\linewidth]{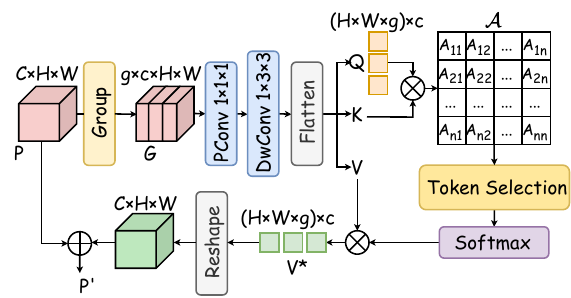}
\caption{The illustration of the proposed Token Selective Fusion Attention (TSFA). The input feature $P$ is first grouped and then fed through a 3D point-wise convolution, followed by a 3D depth-wise convolution to generate the corresponding $\mathbf{Q}\mathbf{K}\mathbf{V}$. Subsequently, a token selection mechanism is utilized for further self-attention operation.}
\label{fig:MSSA}
\end{figure}

Previous research has typically utilized the self-attention mechanism within Transformers to capture long-range dependencies, facilitating the modeling of global spatial or spectral relationships \cite{SpectralFormer,zhang2024deformable}. However, HSI often contains a wide variety of complex land cover types. The conventional Transformer architecture tends to compute attention across all query-key pairs, which inevitably leads to interactions between redundant and irrelevant information, ultimately diminishing interpretative accuracy.  

To overcome this limitation, we propose a novel multi-head Token Selective Fusion Attention (TSFA) mechanism to replace the traditional MHSA mechanism. Furthermore, our TSFA retains the data characteristics of the HSI three-dimensional cube through a grouped operation and employs 3D convolutions to extract tokens with spatial and spectral properties. During the computation of self-attention, the TSFA selectively focuses on the most relevant tokens, effectively disregarding the least relevant information. This approach improves interpretive accuracy while reducing redundant computations.

As illustrated in Fig.~\ref{fig:MSSA}, we first begin by partitioning the input feature $\mathbf{P} \in \mathbb{R}^{C \times H \times W }$ into $g$ groups, each producing a feature map of size $c \times H \times W$.     Then we feed them through a 3D point-wise convolution with a kernel size of $1×1×1$, followed by a 3D depth-wise convolution with a kernel size of $1×3×3$. This process generates the corresponding query $\mathbf{Q}$, key $\mathbf{K}$, and value $\mathbf{V}$. To facilitate the calculation of self-attention, we reshape and transpose the tensors $\mathbf{Q}$, $\mathbf{K}$, and $\mathbf{V}$, adjusting their dimensions by flattening the operation, so each has a shape of $ (H \times W \times g) \times c $. These processes can be extended to the multi-head scenario and expressed as follows:
\begin{equation}
    \mathbf{Q}_i, \mathbf{K}_i, \mathbf{V}_i = Split(\mathcal{F}_{1 \times 3 \times 3}^{dwc}(\mathcal{F}_{1 \times 1 \times 1}(\mathbf{e}_i))),
    \label{eqn:qkv}
\end{equation}
\begin{equation}
    \mathbf{Q}_i, \mathbf{K}_i, \mathbf{V}_i = Flatten(Reshape(\mathbf{Q}_i, \mathbf{K}_i, \mathbf{V}_i)),
    \label{eqn:qkv1}
\end{equation}
where $\mathbf{e}_i$ denotes the feature in $i$th head, $\mathbf{Q}_i, \mathbf{K}_i, \mathbf{V}_i \in \mathbb{R}^{HWg\times c} $ correspond to  $\mathbf{Q}$, $\mathbf{K}$, and $\mathbf{V}$ in $i$th head, respectively, and $g$ represents the number of groups. The $Split(\cdot)$ function refers to the chunk operation performed along the channel dimension. Next, a dense attention matrix $\mathcal{A} \in \mathbb{R}^{HWg \times HWg}$ is generated through the dot product between $\mathbf{Q}_i$ and $\mathbf{K}_i$:

\begin{equation}
\mathcal{A} =  \frac{{\mathbf{Q}_i}{\mathbf{K}_i}^{T} }{\lambda},
    \label{eqn:A}
\end{equation}
where $\lambda = \sqrt{\frac{C}{h}}$, $h$ denotes the number of heads. Following this, the token selection mechanism is applied to identify the top $k$ percent of the most relevant elements within $\mathcal{A}$. For example, with $k=0.8$, the top 80\% of the elements are retained for activation, while the remaining 20\% are masked to 0. To achieve this, a matrix $\mathcal{M}^k$ is created, which operates as follows:
\begin{equation}
\mathcal{M}^k_{pq} =\begin{cases} \mathcal{A}_{pq}, & \mathcal{A}_{pq} > T_k^{[p]} \\
-\infty, & \text{otherwise},
\end{cases} 
    \label{eqn:mask}
\end{equation}
where $\mathcal{M}^k_{pq}$ denotes the value at position $pq$, and $\boldsymbol{T_k} \in \mathbb{R}^{HWg \times 1}$ is the lower threshold vector of the highest $k$ percent attention values in each row of $\mathcal{A}$, i.e, each token has a unique threshold, and $T_k^{[p]}$ is the $p$th value of $\boldsymbol{T_k}$. After that, the selective attention matrix is activated by the softmax function. The activated attention can be written as follows:
\begin{algorithm}[t]
    \caption{Token Selective Fusion Attention}
    \label{alg:TSFA}
    {\bf Input:}
    {Feature $\mathbf{P} \in \mathbb{R}^{C \times H \times W}$, group number $g$, head number $h$, token selection rate $k$.}
    
    \textbf{Initialization:}
    {$\mathbf{e}_{i} \in \mathbb{R}^{d \times H \times W}$ is the $i$-th head of feature, $d=\frac{C}{h}$. $\mathcal{M}^{k} \in \mathbb{R}^{n \times n}$ is initialized with -$\infty$, $n=HWg$. $c=\frac{d}{g}$, $\lambda = \sqrt{d}$.}

\begin{algorithmic}[1]
\FOR{each $\mathbf{e}_i$}
    \STATE $\mathbf{G}_i \leftarrow Group(\mathbf{e}_i)$; \hfill //  $g \times c \times H \times W$
    \STATE $\mathbf{G}_i \leftarrow DWConv(PConv(\mathbf{G}_i))$;  \hfill // $g \times c \times H \times W$
    \STATE $\mathbf{Q}_i, \mathbf{K}_i, \mathbf{V}_i \leftarrow Flatten (Chunk (\mathbf{G}_i))$; \hfill // $ HWg \times c $
    \STATE $\mathcal{A} \leftarrow \frac{\mathbf{Q}_{i}\mathbf{K}_{i}^{T}}{\lambda }$; \hfill // Dense Attention $ HWg \times HWg $
    \STATE $\boldsymbol{T}_k^{[p]} \leftarrow \min(Topk\{\mathcal{A}_{p1}, \cdots,\mathcal{A}_{pn}\}, p=1,\cdots,n)$;\hfill // $1 \times 1$
    \STATE $\mathcal{M}_{pq}^{k} \leftarrow \mathcal{A}_{pq}\; if \; \mathcal{A}_{pq} > T_k^{[p]}$; \hfill // $ HWg \times HWg $
    \STATE $\mathcal{A}^{*} \leftarrow Softmax (\mathcal{M}^{k} )$; \hfill // Selective Attention $ HWg \times HWg $
    \STATE $\mathbf{V}^{*}_{i} \leftarrow \mathcal{A}^{*} \otimes \mathbf{V}_{i}$; \hfill // $ HWg \times c $
    \STATE $\mathbf{V}^{*}_{i} \leftarrow Reshape (\mathbf{V}^{*}_{i}) $; \hfill // $ d \times H \times W$ 
    \STATE $\mathbf{P}^{'}_{i} \leftarrow \mathbf{P}_{i} + \mathbf{V}^{*}_{i}$ \hfill // $ d \times H \times W$ 
    \ENDFOR
\STATE $\mathbf{P}^{'} \leftarrow \mathcal{F}_{1 \times 1}(Concat(\mathbf{P}^{'}_{i})) $ \hfill // $ C \times H \times W$ 
\end{algorithmic}

\end{algorithm}

\begin{equation}
    \mathcal{A}^{*} = Softmax(\mathcal{M}^k),
    \label{eqn:A1}
\end{equation}
Further, the $i$th head selective attention $\mathbf{V}^{*}_{i}$ can be multiplied by $\mathcal{A}^{*}$ and $\mathbf{V}_i$, and then add the input feature $\mathbf{P}_i$ for residual connection:
\begin{equation}
    \mathbf{P}^{'}_{i} = \mathbf{P}_{i} + \mathcal{A}_{i}^{*} \mathbf{V}_{i}.
    \label{eqn:Pi}
\end{equation}

Given our multi-head design, we concatenate the outputs from each head and then aggregate them using a 1×1 convolution, as demonstrated below:
\begin{equation}
    \mathbf{P^{'}}= \mathcal{F}_{1 \times 1}(Concat(\mathbf{P}^{'}_i)).
    \label{eqn:P}
\end{equation}

The detailed implementation of our Token Selective Fusion Attention is presented in Algorithm~\ref{alg:TSFA}.

\section{Experiment}\label{Section4}
In this section, we first provide an overview of the datasets and the implementation details. We then conduct a comprehensive evaluation of the proposed method, including in-depth parameter analysis and module ablation studies. This is followed by a qualitative and quantitative comparison of our approach with current state-of-the-art methods. Finally, we evaluated and discussed the operational efficiency of our model compared to other methods. 

\begin{table}
\caption{Performance contribution of different module on four datasets (reported in OA (\%). Best results are highlighted in \textbf{Bold}. }
\centering
\resizebox{\linewidth}{!}{
\begin{tabular}{ccc|cccc} %
\toprule
Depth & KSFTB & TSFTB       & Pavia University & Houston & Indian Pines & Whu-HongHu \\
\midrule
1&\XSolidBrush & \XSolidBrush  & 87.67 & 92.51 & 85.97 & 88.32\\
9&\XSolidBrush & \XSolidBrush  & 86.33 & 90.04 & 85.99 & 86.58\\
9&\Checkmark & \XSolidBrush    & 93.44 & 96.98 & 92.14 & 92.95\\
9&\XSolidBrush  & \Checkmark     & 95.81 & 97.44 & 93.65 & 93.84\\
9& \Checkmark & \Checkmark  &\textbf{96.59}&\textbf{97.66}&\textbf{95.17}&\textbf{94.59}\\
\bottomrule
\end{tabular}
}
\label{table:ablation}
\end{table}

\begin{table}[!ht]
    \centering
    \caption{The performance of spatial-spectral receptive field selective fusion mechanism on four datasets (reported in OA (\%)). The best results are highlighted in \textbf{Bold}. “w/o Spa/Spe” represents the absence of the fusion of spatial or spectral receptive fields, while “w/ Spa-Spe” indicates that spatial and spectral features of different scales are employed.}
    \resizebox{\linewidth}{!}{
    \begin{tabular}{c c c c c }
    \toprule
       \textbf{DSFormer} &  Pavia University & Houston & Indian Pines & Whu-HongHu\\
        \hline
        \scriptsize{w/o Spa}    &   96.01   &   97.41   &  94.20   &   94.13\\
        \scriptsize{w/o Spe}    &   96.09   &   97.43   &  94.38   &   94.32  \\
        \scriptsize{w/ Spa-Spe}    &   \textbf{96.59}   &   \textbf{97.66}   &  \textbf{95.17}   &   \textbf{94.59}   \\
    \bottomrule
    \end{tabular}
    }
    \label{table:ablation_2}
\end{table}

\subsection{Data Descriptions}

\begin{figure}[t]
\centering
\includegraphics[width=\linewidth]{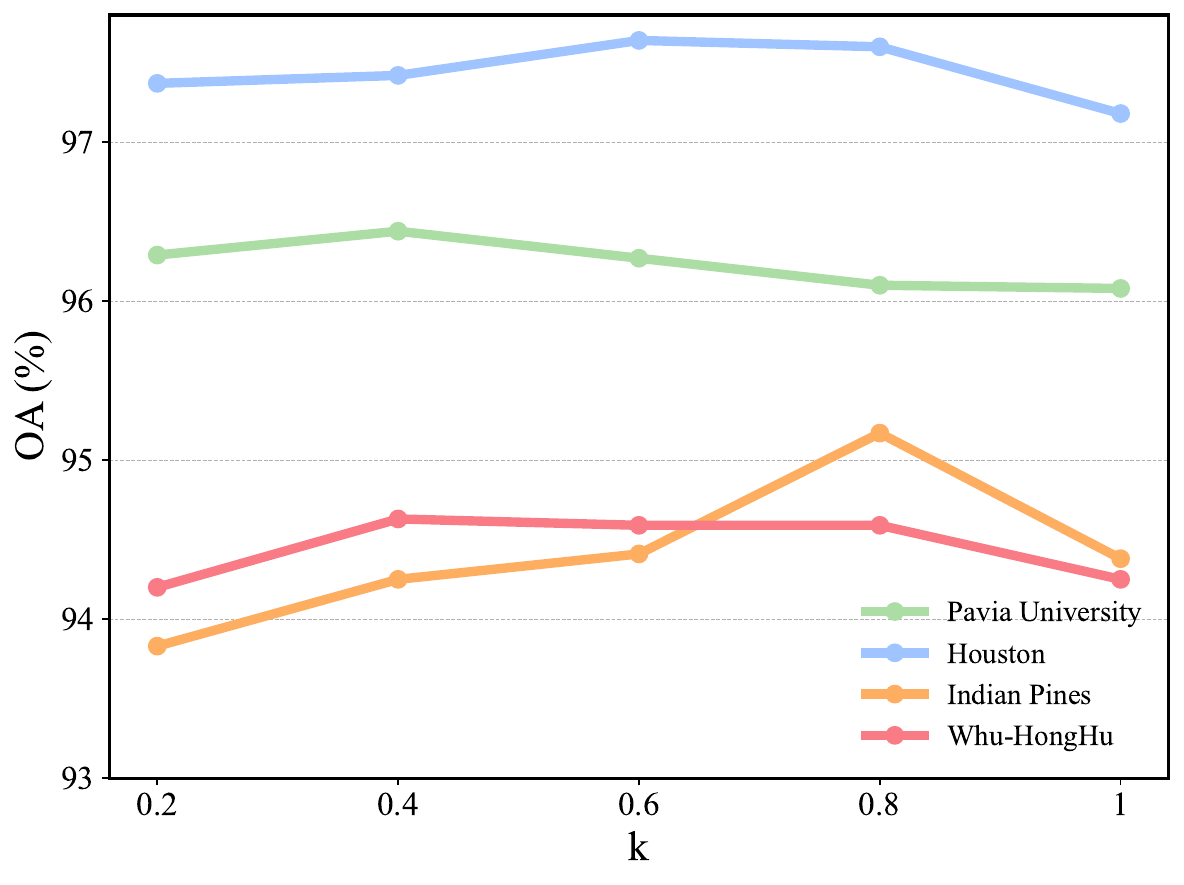}
\caption{Comparison of classification performance across four datasets under different values of $k$.}
\label{fig:Topk}
\end{figure}

\begin{figure}[t]
\centering
\includegraphics[width=\linewidth]{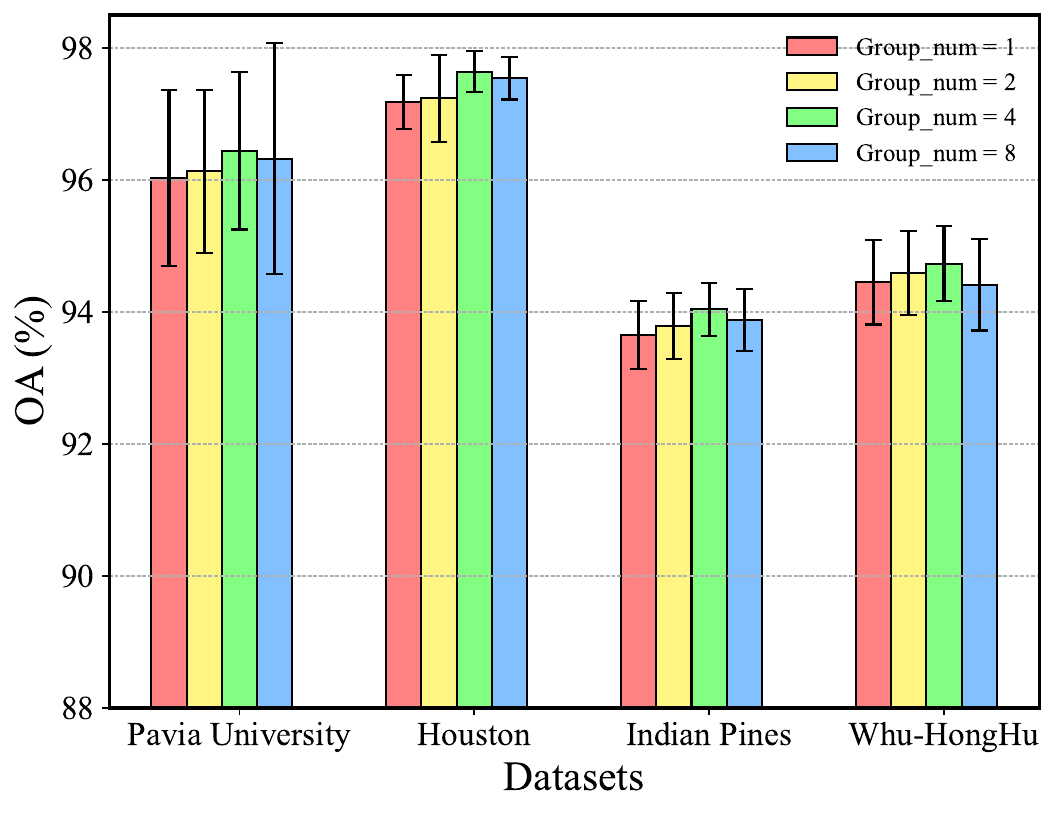}
\caption{Comparison of classification performance across four datasets under different group number $g$.}
\label{fig:group_num}
\end{figure}

\footnotetext[1]{\url{ https://www.ehu.eus/ccwintco/index.php?title=Hyperspectral_Remote_Sensing_Scenes}} 
\begin{enumerate}[(1)]
    \item \textbf{Pavia University\footnotemark[1].} This dataset was acquired by the Reflective Optics System Imaging Spectrometer (ROSIS) on the campus of Pavia University in northern Italy in 2001. It has 103 spectral bands. The dataset consists of 610 by 340 pixels and covers 9 main categories of interest. 
    \item \textbf{Houston.} The Houston dataset was acquired over the University of Houston campus and its surrounding areas in 2012 \citep{houston2013}. It has 144 spectral bands in the wavelength range of 400-1000 nm and a spatial resolution of 2.5 m. It consists of 349 by 1905 pixels and covers 15 categories. 
    \item \textbf{Indian Pines\footnotemark[1].} The AVIRIS sensor acquired this dataset in North-west Indiana in 1992. It has a dimension of 145 by 145 pixels and a spatial resolution of 20m. It covers 200 spectral bands in the wavelength range of 400-2500 nm after removing the water absorption bands. The dataset includes 16 vegetation classes.
    \item \textbf{Whu-HongHu.} This dataset was collected using a UAV platform in HongHu City in 2017 \citep{zhong2020whu}. It comprises 270 spectral bands ranging from 400 to 1000 nm, with a high spatial resolution of approximately 0.04m. The dimensions of the imagery are 940 by 475 pixels, which includes 17 typical crop types. 
\end{enumerate}

\begin{table*}[!htb]
{ 
\centering
\caption{Quantitative classification results of the Pavia University dataset. Best results are shown in \textbf{bold}.}\label{table:PAVIA_OA}
\resizebox{1\textwidth}{!}{
\begin{tabular}{cccccccccccc}
    \toprule
Class      & 3-DCNN      & SSFCN      & SACNet     & FcontNet   & SpectralFormer & SSFTT       & GAHT       & MorphFormer         & GSCViT     & DSFormer             \\
\midrule
1          & 77.43±7.99  & 77.51±1.48 & 80.10±3.43  & 82.17±3.72 & 66.00±11.47    & 77.11±17.13 & 88.39±9.29 & 87.79±6.34          & 93.48±3.62 & \textbf{96.08±1.65} \\
2          & 70.94±10.10 & 78.88±2.21 & 80.12±3.82 & 90.74±2.95 & 77.25±8.09     & 69.92±20.30 & 89.27±4.43 & 90.40±4.98          & 91.69±6.45 & \textbf{95.76±3.15} \\
3          & 68.47±14.58 & 76.25±2.10  & 82.13±4.84 & 95.35±2.63 & 80.32±11.09    & 60.63±14.57 & 87.25±6.23 & 74.95±16.35         & 94.09±4.18 & \textbf{95.57±2.45} \\
4          & 93.48±7.12  & 89.31±1.52 & 77.97±3.18 & 86.59±4.38 & 91.10±2.12     & 97.26±2.03  & 96.87±1.65 & 93.50±3.43          & 93.61±4.88 & \textbf{97.09±0.76} \\
5          & 99.56±0.42  & 98.09±1.53 & 95.70±1.47  & 99.64±0.36 & 99.86±0.21     & 99.85±0.22  & 99.85±0.19 & 99.54±0.51          & 99.86±0.15 & \textbf{99.94±0.12} \\
6          & 75.85±13.08 & 82.65±2.37 & 91.96±2.09 & 96.54±2.14 & 77.09±11.96    & 77.10±16.86 & 92.96±5.98 & 94.20±4.21          & 93.86±4.74 & \textbf{98.00±1.35} \\
7          & 88.00±3.75  & 86.72±1.84 & 92.35±2.37 & 95.45±0.89 & 75.66±10.84    & 78.88±20.94 & 97.07±3.82 & 93.50±5.26          & 98.53±1.17 & \textbf{99.51±0.42} \\
8          & 82.22±15.26 & 84.50±2.23  & 86.65±1.12 & 90.72±3.21 & 62.86±20.38    & 76.74±20.03 & 88.08±5.43 & 91.89±6.18          & 95.12±1.94 & \textbf{97.04±0.85} \\
9          & 99.42±0.67  & 99.09±0.63 & 98.46±0.77 & 95.53±4.20  & 96.80±1.23     & 98.83±1.95  & 99.35±0.49 & \textbf{99.75±0.25} & 99.64±0.44 & 99.12±0.66          \\ \hline
OA (\%)    & 77.01±4.48  & 81.48±1.45 & 83.27±1.72 & 90.54±2.17 & 76.46±2.36     & 75.79±9.10  & 90.69±2.44 & 90.62±2.29          & 93.40±2.43 & \textbf{96.59±1.41} \\
$\kappa$ (\%) & 70.95±5.22  & 76.23±1.78 & 78.53±2.03 & 87.68±2.74 & 69.94±2.64     & 69.92±9.98  & 87.91±3.09 & 87.77±2.89          & 91.39±3.05 & \textbf{95.52±1.82} \\
AA (\%)    & 83.93±2.36  & 85.89±0.98 & 87.27±0.91 & 92.53±1.66 & 80.77±1.91     & 81.81±4.10  & 93.23±1.38 & 91.72±1.69          & 95.54±1.07 & \textbf{97.57±0.51}\\   
\bottomrule
\end{tabular}
}}
\end{table*}

\begin{figure*}[h]
\centering
\includegraphics[width=\linewidth]{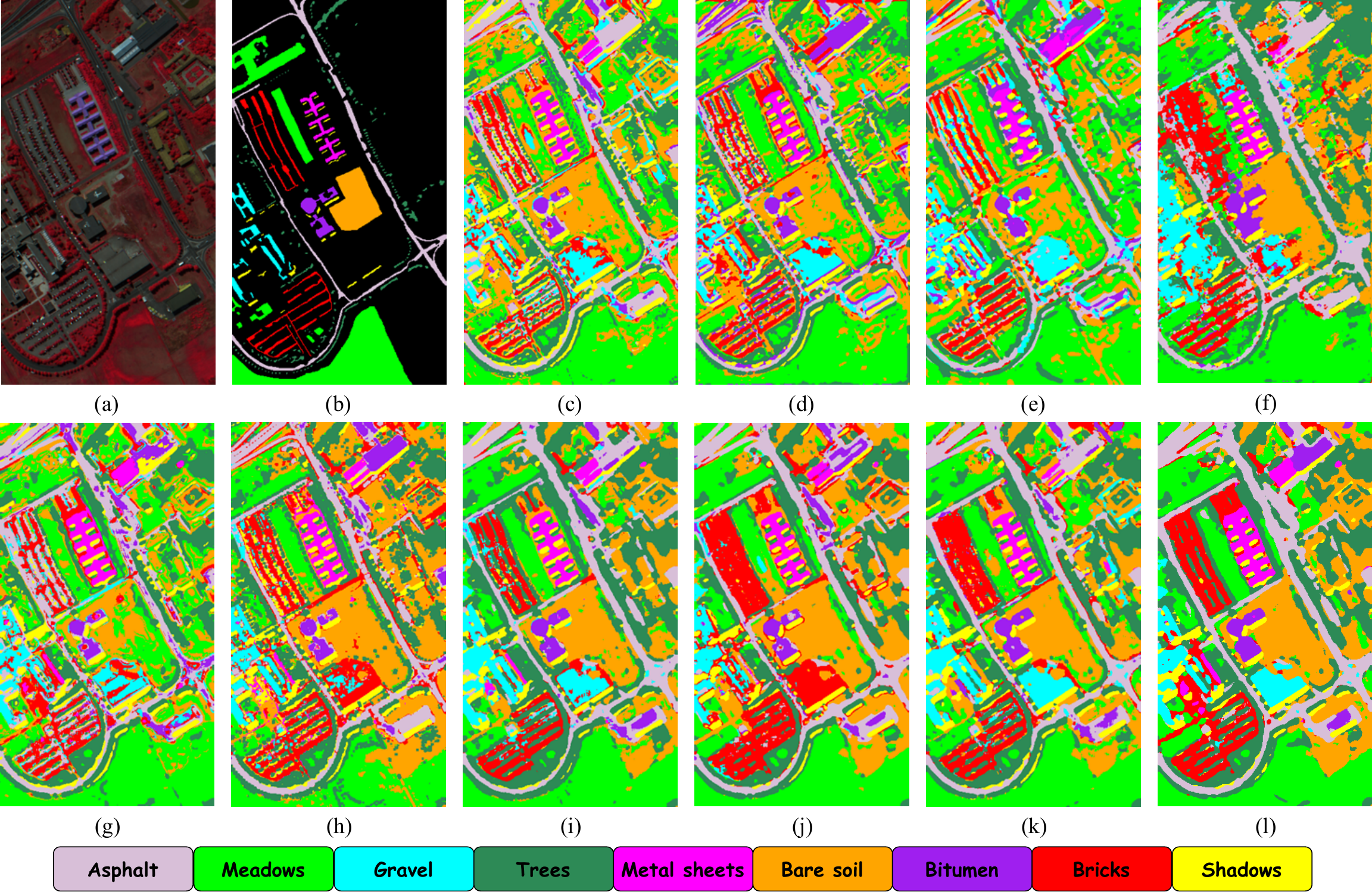}
\caption{ Visualization of the classification maps on the Pavia University dataset.   (a) False-color image. (b) Ground truth. (c) 3-DCNN. (d) SSFCN. (e) SACNet. (f) FcontNet. (g) SpectralFormer. (h) SSFTT. (i) GAHT. (j) MorphFormer. (k) GSCViT. (l) DSFormer.}
\label{fig:PaviaMap}
\end{figure*}

\subsection{Experimental Settings}

The experiments in this study were conducted on the PyTorch platform, utilizing an NVIDIA GeForce RTX 3090 GPU. For the Pavia University dataset, 30 samples per class were randomly selected for training. For the other three datasets, 50 samples per class were chosen, except for classes 1, 7, and 9 of the Indian Pines dataset, where only 15 samples were selected. The remaining labeled samples were used for testing. 
The training process used the AdamW optimizer, starting with an initial learning rate of $1e-4$ and a weight decay of $1e-5$, was run for 500 epochs, extract patch cubes of size $C \times10\times 10$, patch size in transformer is 2, and embedding dimension is fixed with 128 in all datasets. To quantitatively assess classification performance, we used three widely recognized evaluation metrics: overall accuracy (OA), average accuracy (AA), and kappa coefficient ($\kappa$). Each experiment was repeated 10 times with different random seeds, and the classification results are presented as mean and standard deviation.

\begin{table*}[!htb]
{ 
\centering
\caption{Quantitative classification results of the Houston dataset. Best results are shown in \textbf{bold}.}\label{table:Houston_OA}
\resizebox{1\textwidth}{!}{
\begin{tabular}{cccccccccccc}
    \toprule
Class       & 3-DCNN     & SSFCN        & SACNet       & FcontNet      & SpectralFormer & SSFTT       & GAHT                & MorphFormer & GSCViT              & DSFormer          \\
\midrule
1          & 91.96±6.02 & 94.17 ± 1.94 & 87.38 ± 1.43 & 94.44 ± 1.57 & 94.32±3.37 & 87.76±5.05  & 93.61±4.89   & 96.79±1.73 & 96.81±1.95          & \textbf{98.25±1.26} \\
2          & 92.57±6.08 & 96.80 ± 2.54  & 79.50 ± 3.08  & 90.35 ± 1.89 & 96.45±1.04 & 94.97±5.43  & 97.52±3.83   & 97.57±3.50 & 99.31±0.86          & \textbf{99.37±0.34} \\
3          & 99.21±0.95 & 99.09 ± 0.49 & 98.56 ± 0.87 & 99.41 ± 0.64 & 99.43±0.42 & 90.36±17.37 & 99.57±0.49   & 99.32±0.90 & 99.58±0.54          & \textbf{99.61±0.26} \\
4          & 93.58±2.12 & 89.61 ± 1.12 & 82.05 ± 3.02 & 90.59 ± 2.89 & 93.72±2.53 & 97.85±2.74  & 96.81±2.57   & 96.73±2.60 & 99.39±0.65          & \textbf{99.56±0.67} \\
5          & 98.98±0.83 & 97.23 ± 0.58 & 82.10 ± 3.33  & 97.87 ± 0.80  & 98.70±0.92 & 98.72±1.85  & 99.66±0.57   & 99.35±0.82 & \textbf{100} & 99.87±0.24          \\
6          & 92.95±5.28 & 91.78 ± 1.09 & 89.20 ± 2.17  & 95.85 ± 1.68 & 96.33±1.87 & 97.09±3.11  & 98.95±1.18   & 98.87±1.19 & \textbf{99.96±0.11 }         & 99.78±0.65 \\
7          & 87.83±2.82 & 82.31 ± 1.36 & 73.04 ± 1.83 & 85.48 ± 2.66 & 78.10±4.61 & 87.73±6.38  & 94.11±2.77   & 93.45±2.06 & 95.48±2.67          & \textbf{97.95±1.06} \\
8          & 72.97±6.39 & 83.15 ± 2.30  & 62.19 ± 2.66 & 80.45 ± 3.09 & 84.86±3.65 & 78.49±5.71  & 89.36±2.98   & 87.24±4.22 & 89.66±4.75          & \textbf{92.28±2.94} \\
9          & 83.60±4.00 & 81.41 ± 1.20  & 68.82 ± 2.52 & 80.05 ± 2.10  & 76.15±4.90 & 89.54±5.53  & 90.87±3.73   & 90.52±2.48 & 90.87±4.30          & \textbf{94.38±2.07} \\
10         & 85.53±5.68 & 88.07 ± 1.97 & 90.93 ± 2.32 & 92.56 ± 3.15 & 89.97±3.90 & 63.59±27.82 & 93.02±5.28   & 96.29±2.30 & \textbf{98.09±1.25}          & 97.69±3.11 \\
11         & 71.20±4.31 & 84.08 ± 2.49 & 84.14 ± 1.93 & 94.36 ± 2.81 & 76.36±3.41 & 82.01±14.44 & 94.18±2.62   & 93.79±2.55 & 96.25±2.02          & \textbf{98.03±1.81} \\
12         & 84.33±4.93 & 79.37 ± 2.80  & 84.57 ± 2.84 & 94.61 ± 1.99 & 80.35±4.81 & 67.31±29.28 & 92.53±3.96   & 94.22±3.13 & 93.44±4.63 & \textbf{95.16±2.97}          \\
13         & 94.84±0.66 & 88.76 ± 1.74 & 82.98 ± 3.61 & 97.28 ± 1.33 & 76.11±4.69 & 96.56±2.74  & 97.18±1.48   & 98.42±1.60 & 98.59±1.65          & \textbf{99.38±0.66} \\
14         & 98.68±1.31 & 98.39 ± 0.48 & 98.92 ± 0.69 & \textbf{100}      & 96.85±1.76 & 99.79±0.39  & \textbf{100} & 99.81±0.56 & \textbf{100}        & 99.97±0.08          \\
15         & 99.64±0.39 & 95.77 ± 0.60  & 94.23 ± 2.15 & 99.64 ± 0.32 & 98.84±1.08 & 99.82±0.21  & 99.89±0.18   & 99.69±0.42 & 99.66±0.42          & \textbf{100}        \\ \hline
OA (\%)    & 88.14±1.32 & 88.89±1.00   & 81.74±0.93   & 91.47±1.32   & 88.10±0.74 & 86.72±4.09  & 94.99±0.91   & 95.36±0.99 & 96.52±0.60          & \textbf{97.66±0.51} \\
$\kappa$ (\%) & 87.17±1.43 & 87.99±1.08   & 80.26±1.00   & 90.78±1.42   & 87.14±0.80 & 85.65±4.41  & 94.58±0.99   & 94.98±1.07 & 96.23±0.65          & \textbf{97.47±0.56} \\
AA (\%)    & 89.86±1.18 & 90.00±0.85   & 83.91±0.90   & 92.86±1.07   & 89.10±0.68 & 88.77±3.57  & 95.82±0.81   & 96.14±0.88 & 97.14±0.49          & \textbf{98.09±0.42}\\        
\bottomrule
\end{tabular}
}}
\end{table*}

\begin{figure*}[!htb]
\centering
\includegraphics[width=\linewidth]{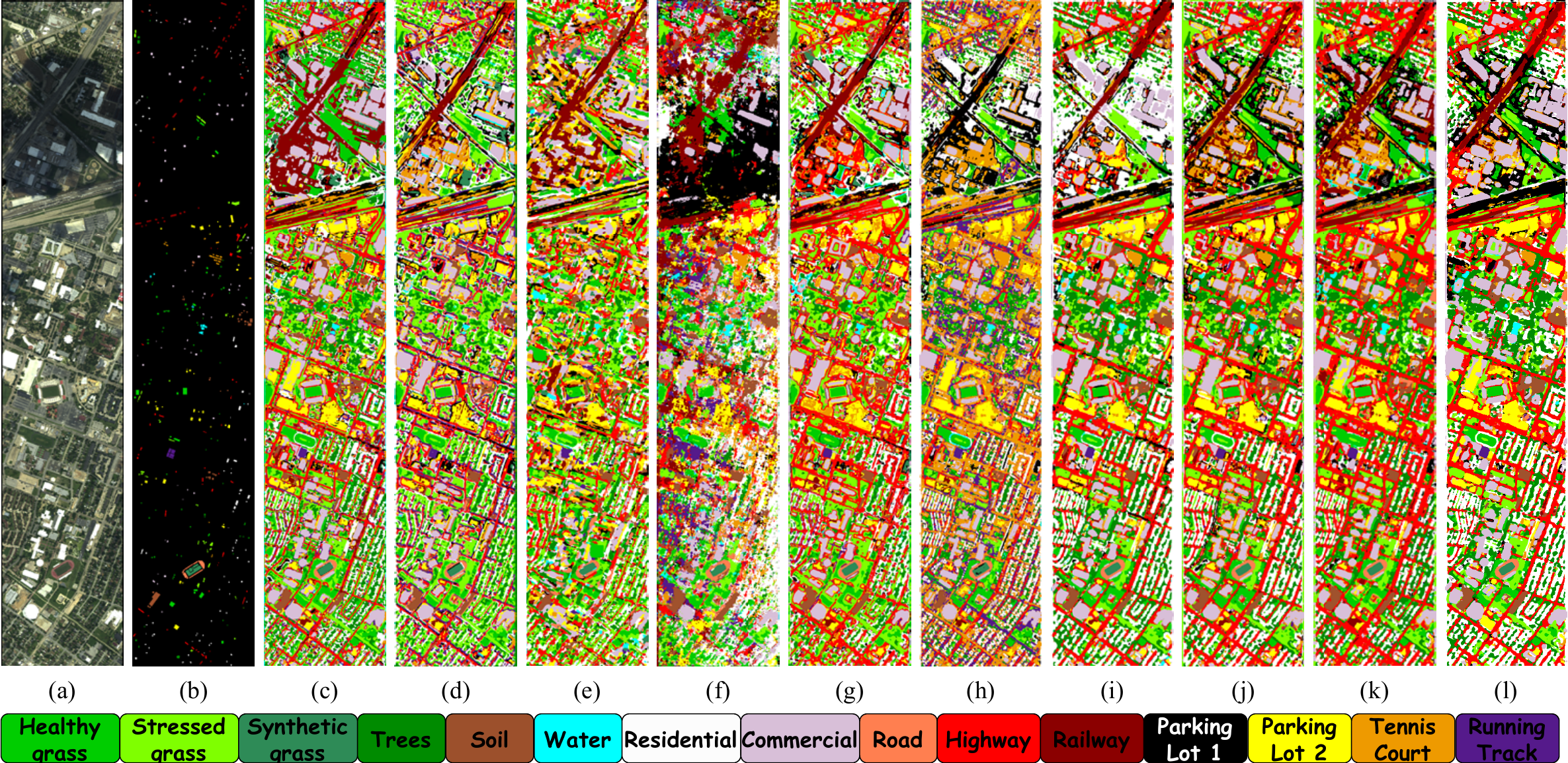}
\caption{ Visualization of the classification maps on the Houston dataset.   (a) False-color image. (b) Ground truth. (c) 3-DCNN. (d) SSFCN. (e) SACNet. (f) FcontNet. (g) SpectralFormer. (h) SSFTT. (i) GAHT. (j) MorphFormer. (k) GSCViT. (l) DSFormer.}
\label{fig:HoustonMap}
\end{figure*}

\subsection{Parameter Analysis}

\textbf{(1) Effect of Different Values $k$.} In the TSFTB module, one of the core parameters to select the $k$ tokens is the value of $k$. To identify the optimal value of $k$, we performed four experiments with $k$ set to 0.2, 0.4, 0.6, 0.8, and 1, respectively. When $k$ is set to 0.8, the TSFTB module selects the most relevant 80\% tokens for the spatial-spectral context modeling, while the remaining 20\% are considered the least relevant and are therefore excluded from the information interaction process. When $k$ is set to 1, the attention mechanism in the TSFTB module degenerates into the conventional full self-attention form. Fig.~\ref{fig:Topk} presents the OA results in four datasets with different values of $k$. The experimental results show that when $k$ is set to 0.2, the classification accuracy is relatively low due to the insufficient information available for long-range modeling. As the value of $k$ increases, the OA improves, indicating that more pixels contribute to the fusion of contextual relationships, enabling the network to capture richer and more effective information for the classification task. However, when $k$ exceeds the optimal value, OA begins to decline, likely due to the inclusion of irrelevant or redundant information, 
which negatively impacts classification performance. Finally, based on the experimental results, we selected $k$ values of 0.4, 0.6, 0.8, and 0.8 across the four datasets to achieve optimal classification performance. Furthermore, we observe that performance in most $k$ configurations outperforms the traditional self-attention mechanism ($k=1$), validating the effectiveness of the token selection strategy.

\textbf{(2) Effect of Different Group Numbers $g$.} A critical parameter in the DSFormer architecture is the number of groups, denoted as $g$. To determine the optimal value for $g$, we conducted four experimental configurations with $g$ set to 1, 2, 4, and 8, respectively. By utilizing grouping and 3D convolutions, the TSFTB module preserves the spatial-spectral characteristics of HSIs, enabling the extraction of tokens that incorporate both spatial and spectral features. The value of $g$ directly impacts the size of the attention matrix, with larger $g$ values leading to increased computational complexity due to the expansion of the matrix. When $g$ is set to 1, the TSFTB module reduces to purely spatial token selection without any grouping. The experimental results, as illustrated in Fig.~\ref{fig:group_num}, show that when $g$ is set to 1, no grouping strategy is applied to preserve spatial-spectral characteristics, resulting in a relatively lower classification accuracy across the four datasets. In contrast, the grouping strategy significantly improves classification performance. Notably, the optimal classification performance is achieved with $g$ is set to 4 across all four datasets, while maintaining relatively efficient computational complexity.

\begin{table*}[!htb]
{ 
\centering
\caption{Quantitative classification results of the Indian Pines dataset. Best results are shown in \textbf{Bold}.}\label{table:Indian_OA}
\resizebox{1\textwidth}{!}{
\begin{tabular}{cccccccccccc}
    \toprule
Class       & 3-DCNN       & SSFCN               & SACNet       & FcontNet              & SpectralFormer & SSFTT        & GAHT                & MorphFormer         & GSCViT              & DSFormer                \\
\midrule
1          & 93.87±2.26   & 93.23 ± 2.26 & 99.68 ± 0.97 & 90.32 ± 6.12 & 90.00±7.83 & 96.77±4.33   & \textbf{100} & 97.10±3.04          & 99.03±1.48          & 99.35±1.29          \\
2          & 60.45±9.78   & 75.64 ± 1.76 & 89.33 ± 1.66 & 80.86 ± 6.15 & 68.72±3.71 & 84.45±7.24   & 88.08±3.69   & 88.15±2.65          & 89.47±2.85          & \textbf{93.22±2.14} \\
3          & 70.64±4.44   & 83.36 ± 2.00 & 93.59 ± 1.95 & 94.21 ± 2.93 & 82.14±6.69 & 79.05±26.67  & 93.76±3.85   & 94.05±5.19          & \textbf{96.53±2.67} & 96.08±2.40          \\
4          & 95.72±3.80   & 92.51 ± 3.08 & 97.86 ± 1.29 & 95.29 ± 1.24 & 94.60±2.40 & 97.65±2.74   & 99.63±0.79   & \textbf{99.89±0.32} & \textbf{99.89±0.21} & 99.79±0.43          \\
5          & 89.49±2.49   & 90.30 ± 3.39 & 97.76 ± 0.75 & 94.83 ± 3.19 & 90.81±2.73 & 95.38±2.33   & 96.56±2.07   & 96.95±1.59          & 97.23±1.69          & \textbf{98.48±1.45} \\
6          & 96.85±0.64   & 89.75 ± 1.65 & 98.65 ± 0.79 & 96.74 ± 2.12 & 95.32±1.80 & 96.82±5.89   & 99.34±0.42   & 98.63±1.28          & 99.43±0.61          & \textbf{99.68±0.22} \\
7          & 96.92±3.77   & 98.46 ± 3.08 & \textbf{100} & \textbf{100} & 98.46±4.62 & \textbf{100} & \textbf{100} & \textbf{100}        & \textbf{100}        & \textbf{100}        \\
8          & 98.06±1.90   & 94.35 ± 1.40 & 98.62 ± 0.82 & 99.77 ± 0.26 & 97.27±2.36 & 99.58±0.55   & 99.86±0.28   & 99.74±0.29          & 99.93±0.21          & \textbf{100}        \\
9          & \textbf{100} & \textbf{100} & \textbf{100} & \textbf{100} & 98.00±6.00 & \textbf{100} & \textbf{100} & \textbf{100}        & \textbf{100}        & \textbf{100}        \\
10         & 76.80±6.72   & 75.42 ± 2.65 & 89.34 ± 2.62 & 93.67 ± 2.42 & 81.40±3.86 & 88.37±5.57   & 91.91±2.68   & 91.34±3.36          & \textbf{94.03±2.51} & \textbf{94.03±2.55} \\
11         & 66.88±7.56   & 64.51 ± 3.09 & 91.06 ± 1.24 & 89.84 ± 5.82 & 73.53±4.42 & 60.80±25.84  & 82.13±4.48   & 87.42±4.85          & 89.50±3.21          & \textbf{90.25±1.53} \\
12         & 82.71±7.42   & 80.41 ± 5.46 & 86.10 ± 1.31  & 91.20 ± 2.78  & 74.84±6.65 & 91.14±3.34   & 94.97±2.23   & 95.49±1.51          & 95.82±1.88          & \textbf{96.24±1.73} \\
13         & 99.87±0.26   & 98.39 ± 1.09 & 99.61 ± 0.43 & \textbf{100} & 99.42±0.54 & 99.94±0.19   & \textbf{100} & 99.94±0.19          & 99.94±0.19          & \textbf{100}        \\
14         & 92.24±2.85   & 89.34 ± 1.18 & 96.02 ± 1.04 & 95.74 ± 1.77 & 92.10±2.82 & 96.29±2.00   & 95.62±1.83   & 97.42±1.54          & 97.77±0.88          & \textbf{98.52±0.66} \\
15         & 94.85±4.19   & 90.65 ± 2.18 & 99.97 ± 0.09 & 99.52 ± 0.40  & 93.45±4.24 & 97.11±1.79   & 98.72±1.20   & 98.72±1.40          & 99.38±0.80          & \textbf{99.88±0.15} \\
16         & \textbf{100} & 99.30 ± 1.07 & \textbf{100} & 99.07 ± 1.14 & 99.53±0.93 & 99.77±0.70   & \textbf{100} & 99.77±0.70          & \textbf{100}        & \textbf{100}        \\ \hline
OA (\%)    & 78.28±2.01   & 79.40±1.30   & 93.05±0.60   & 91.93±3.22   & 81.88±1.36 & 83.47±9.07   & 91.39±1.32   & 92.91±1.52          & 94.26±0.89          & \textbf{95.17±0.93} \\
$\kappa$ (\%) & 75.30±2.26   & 76.68±1.45   & 92.04±0.68   & 90.78±3.66   & 79.43±1.51 & 81.43±9.94   & 90.17±1.48   & 91.89±1.73          & 93.43±1.01          & \textbf{94.47±1.07} \\
AA (\%)    & 88.46±1.28   & 88.48±1.09   & 96.10±0.35   & 95.06±1.91   & 89.35±1.10 & 92.70±3.71   & 96.29±0.45   & 96.54±0.82          & 97.37±0.39          & \textbf{97.85±0.44}\\        
\bottomrule
\end{tabular}
}}
\end{table*}

\begin{figure*}[!htb]
\centering
\includegraphics[width=\linewidth]{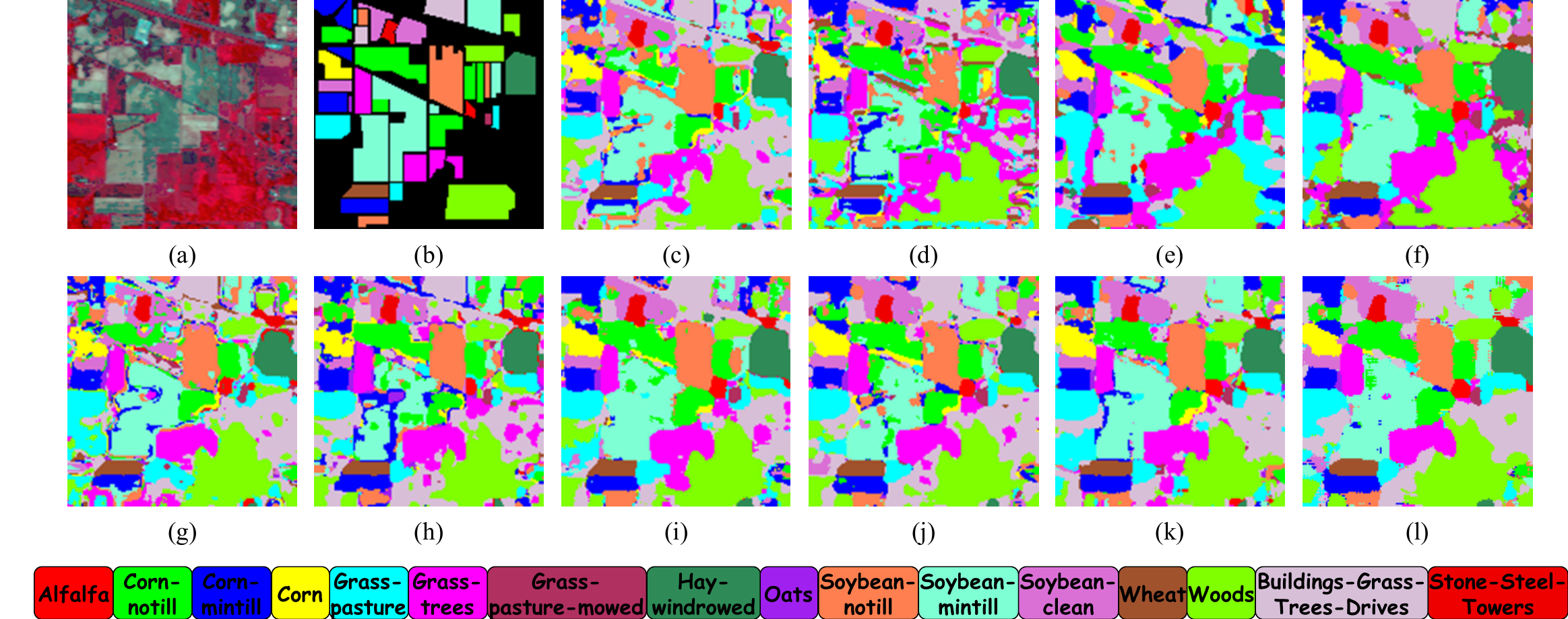}
\caption{ Visualization of the classification maps on the Indian Pines dataset.   (a) False-color image. (b) Ground truth. (c) 3-DCNN. (d) SSFCN. (e) SACNet. (f) FcontNet. (g) SpectralFormer. (h) SSFTT. (i) GAHT. (j) MorphFormer. (k) GSCViT. (l) DSFormer.}
\label{fig:IndianMap}
\end{figure*}

\subsection{Ablation Study}
\textbf{(1) Performance Contribution of Each Module.} We explore the impact of each module within DSFormer on classification performance through an ablation study, as shown in Table~\ref{table:ablation}. The baseline architecture utilizes a single $3 \times 3$ convolutional layer, as well as a variant that employs nine consecutively stacked $3 \times 3$ convolutional layers to align with the depth of the DSFormer network. The results presented indicate that both the KSFTB and TSFTB modules significantly enhance the network's classification performance. For example, on the Pavia University dataset, the baseline with single convolutional layer achieves an OA of only 87.67\%. However, increasing the number of network layers did not lead to improved accuracy. In fact, employing nine convolutional layers yielded an OA of only 86.33\%. The introduction of the KSFTB module improved the OA to 93.44\%, while the application of the TSFTB module increased the OA to 95.81\%. Each of these modules contributes to improved classification performance through their fusion mechanisms. Specifically, the KSFTB module improves recognition accuracy for objects of varying scales by enabling the network to adaptively select and fuse appropriate receptive field features from spatial and spectral respectives. While the TSFTB module enhances classification by computing self-attention spatial spectral contexts on the most relevant tokens. 
This process reduces redundancy and interference from irrelevant information, thus improving the integration and utilization of useful features. When both KSFTB and TSFTB modules are applied together, the classification performance is further improved, with the OA yielding 96.59\%. In summary, the results of the ablation study have demonstrated that KSFTB and TSFTB modules are able to simultaneously improve the network's classification performance on all four HSI datasets, while the combination of two modules exhibits a synergistic effect, achieving the best OAs and confirming the effectiveness of these modules in enhancing network classification performance.

\begin{table*}[!htb]
{\centering
\caption{Quantitative classification results of the Whu-HongHu dataset. Best results are shown in \textbf{bold}.}\label{table:HongHu_OA}
\resizebox{1\textwidth}{!}{
\begin{tabular}{cccccccccccc}
    \toprule
Class       & 3-DCNN      & SSFCN               & SACNet       & FcontNet              & SpectralFormer & SSFTT       & GAHT                & MorphFormer         & GSCViT              & DSFormer\\
\midrule
1          & 90.78±2.65 & 92.34 ± 1.76 & 88.22 ± 1.86 & 89.94 ± 2.38          & 95.28±1.04 & 82.37±15.01 & 96.13±0.81          & 94.97±4.13          & 93.46±5.86          & \textbf{96.35±0.77} \\
2          & 92.10±1.50 & 88.75 ± 1.34 & 89.69 ± 1.26 & 88.54 ± 5.40          & 90.01±2.12 & 69.44±31.34 & 96.23±1.47 & 95.61±2.81          & 95.07±4.31          & \textbf{96.84±1.23}          \\
3          & 80.27±7.44 & 81.16 ± 0.94 & 85.64 ± 1.10 & 85.93 ± 0.73          & 78.32±1.68 & 80.07±11.96 & 84.00±8.94          & 87.78±4.15          & 89.40±2.69          & \textbf{92.66±1.20} \\
4          & 90.10±2.58 & 87.93 ± 1.28 & 91.50 ± 1.90 & 89.94 ± 5.18          & 93.20±2.40 & 84.41±11.17 & 95.46±0.70          & 93.31±2.80          & 95.56±5.62          & \textbf{97.09±1.01} \\
5          & 85.49±3.52 & 84.99 ± 1.04 & 90.95 ± 0.75 & 93.46 ± 5.21          & 85.74±5.17 & 72.78±19.67 & 93.45±2.38          & 94.44±3.14          & 90.68±9.77         & \textbf{96.76±1.79} \\
6          & 90.33±4.36 & 85.74 ± 1.27 & 86.64 ± 1.04 & 91.05 ± 2.85          & 90.41±2.92 & 66.43±29.85 & 94.12±2.52          & 94.68±2.32          & 93.32±2.90          & \textbf{94.95±1.93} \\
7          & 66.40±3.44 & 61.18 ± 2.27 & 57.62 ± 2.61 & 76.95 ± 3.22          & 70.49±3.32 & 69.35±14.28 & 78.60±3.92          & 78.57±3.69          & 80.01±5.05          & \textbf{84.44±1.99}        \\
8          & 59.23±6.12 & 51.81 ± 2.87 & 81.22 ± 3.34 & 97.54 ± 1.79          & 63.43±3.11 & 64.45±12.07 & 80.02±4.75          & 84.93±7.33          & 83.12±3.24          & \textbf{92.69±2.15}        \\
9          & 95.45±1.16 & 95.04 ± 0.74 & 87.93 ± 1.76 & 92.61 ± 1.73          & 96.94±0.86 & 95.58±3.40  & 97.27±1.74          & 96.12±2.53          & \textbf{97.55±1.37} & 97.21±1.80        \\
10         & 82.54±4.28 & 66.45 ± 2.86 & 59.82 ± 2.38 & 86.82 ± 2.46          & 61.54±5.74 & 81.11±8.85  & 89.57±2.24          & 88.08±5.46          & 82.34±11.69          & \textbf{93.54±1.42} \\
11         & 68.36±5.18 & 56.68 ± 2.51 & 57.55 ± 3.51 & 80.69 ± 4.14          & 71.51±3.36 & 55.64±19.38 & 85.85±3.57          & 87.15±4.04          & 87.65±3.80         & \textbf{92.58±1.61}    \\
12         & 70.88±5.60 & 58.97 ± 2.42 & 78.58 ± 3.28 & 89.90 ± 3.60          & 70.96±4.17 & 56.33±20.32 & 79.35±5.21          & 82.53±5.54          & 81.50±9.10          & \textbf{88.52±2.96} \\
13         & 66.08±7.33 & 54.73 ± 1.88 & 67.93 ± 2.28 & 86.36 ± 4.36          & 67.83±5.20 & 67.83±5.20  & 78.49±4.74          & 80.09±4.50          & 78.77±4.33          & \textbf{84.46±2.80}        \\
14         & 86.10±5.35 & 76.66 ± 2.60 & 78.39 ± 1.78 & 95.91 ± 2.47          & 85.54±3.17 & 85.43±8.74  & 94.55±2.26          & 95.28±3.66          & 92.92±5.11          & \textbf{97.54±0.91} \\
15         & 94.50±1.95 & 93.32 ± 1.00 & 98.58 ± 0.83 & 99.35 ± 0.47          & 94.02±2.90 & 97.13±2.83  & 98.26±1.01          & 98.94±1.06          & 98.89±0.66          & \textbf{99.26±0.76} \\
16         & 91.36±3.75 & 89.41 ± 1.53 & 90.74 ± 2.92 & 94.29 ± 2.02          & 89.36±3.51 & 91.33±3.20  & 94.82±1.69          & 94.05±4.07          & \textbf{98.31±1.48}          & 97.26±1.37       \\
17         & 91.41±3.83 & 89.81 ± 2.16 & 95.75 ± 1.78 & 95.14 ± 2.61          & 89.21±2.37 & 92.08±3.87  & 96.65±1.96          & 97.54±1.77 & 95.30±5.49          & \textbf{97.66±1.49}         \\
18         & 91.13±2.78 & 90.33 ± 2.19 & 94.03 ± 0.81 & 97.23 ± 1.67          & 90.66±3.82 & 64.21±25.62 & 95.68±2.52          & 98.21±1.22 & 96.14±2.73 & \textbf{98.80±0.90}          \\
19         & 89.90±2.60 & 85.11 ± 1.81 & 81.46 ± 1.40 & 86.18 ± 2.97          & 90.03±1.94 & 67.38±24.16 & 94.89±0.94          & 94.36±1.48          & 94.21±2.12          & \textbf{95.80±0.74}   \\
20         & 92.40±2.76 & 96.38 ± 0.78 & 91.49 ± 2.31 & \textbf{99.56 ± 0.16} & 94.45±2.00 & 75.31±22.05 & 95.69±2.37          & 96.27±2.90          & 97.42±1.69          & 98.90±0.52          \\
21         & 83.04±9.61 & 87.71 ± 1.64 & 95.23 ± 2.22 & \textbf{99.69 ± 0.20} & 87.86±5.67 & 55.38±36.65 & 93.64±12.16         & 92.61±15.09         & 89.48±11.80         & 99.01±0.66          \\
22         & 90.53±4.40 & 87.92 ± 1.82 & 94.77 ± 3.57 & \textbf{99.13 ± 0.66} & 91.54±3.50 & 88.84±9.76  & 95.71±2.00          & 97.01±2.54          & 97.14±1.80          & 97.95±1.12          \\ \hline
OA (\%)    & 85.16±1.18 & 81.26±0.70   & 84.12±1.13   & 89.10±2.83            & 86.36±1.05 & 77.17±9.41  & 91.65±0.85          & 91.19±1.52          & 91.79±2.32         & \textbf{94.59±0.50} \\
$\kappa$ (\%) & 81.67±1.39 & 76.98±0.81   & 80.33±1.31   & 86.55±3.32            & 83.03±1.24 & 72.46±10.69 & 89.55±1.05          & 89.04±1.82          & 89.76±2.72         & \textbf{93.20±0.62} \\
AA (\%)    & 84.02±1.09 & 80.11±0.72   & 83.81±0.68   & 91.65±1.31            & 84.01±0.83 & 75.03±9.24  & 91.29±0.90          & 91.93±1.60          & 91.28±1.23          & \textbf{95.01±0.21}\\     
\bottomrule
\end{tabular}
}}
\end{table*}

\begin{figure*}[!htb]
\centering
\includegraphics[width=\linewidth]{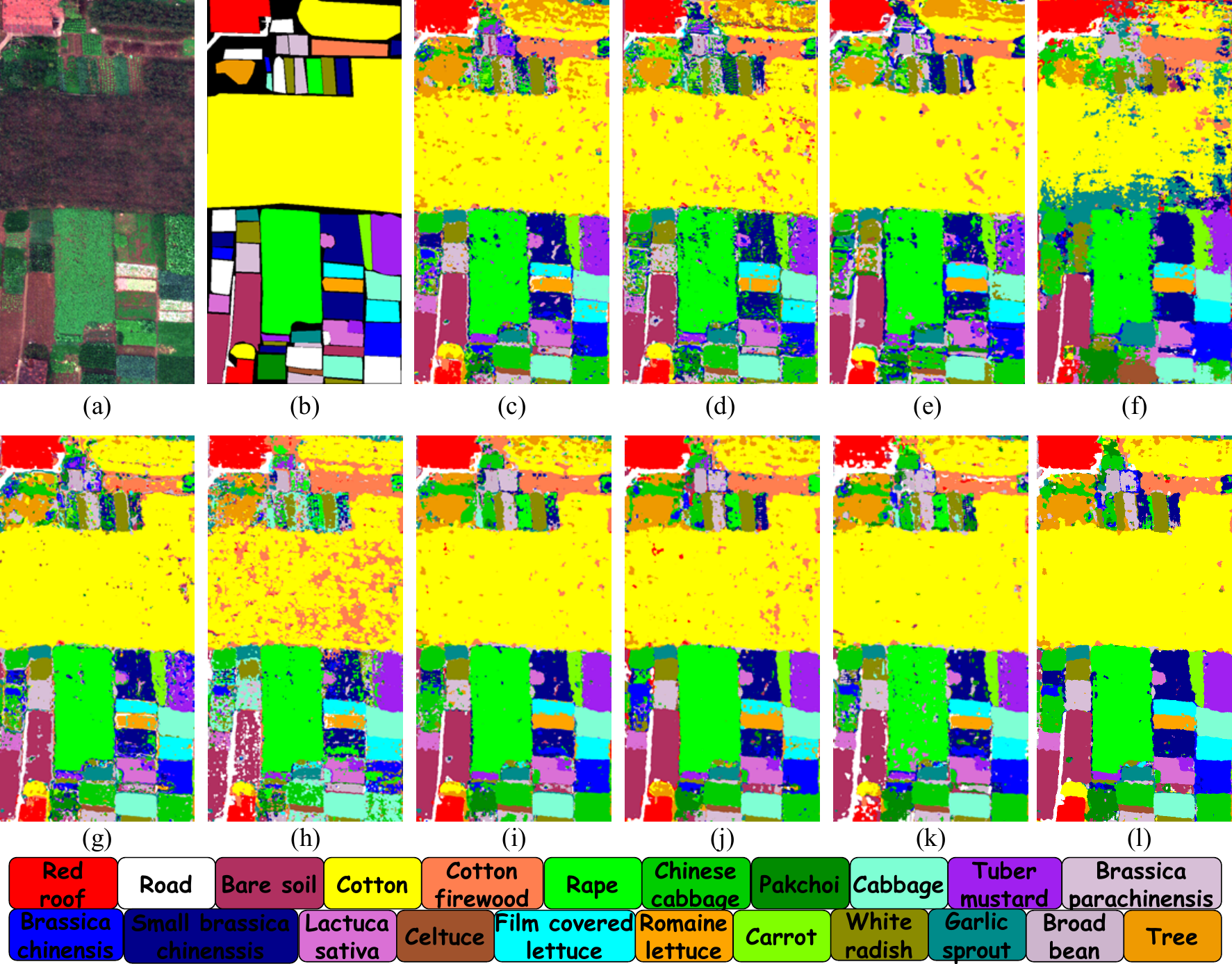}
\caption{ Visualization of the classification maps on the Whu-HongHu dataset.   (a) False-color image. (b) Ground truth. (c) 3-DCNN. (d) SSFCN. (e) SACNet. (f) FcontNet. (g) SpectralFormer. (h) SSFTT. (i) GAHT. (j) MorphFormer. (k) GSCViT. (l) DSFormer.}
\label{fig:HongHuMap}
\end{figure*}

\begin{figure*}[!htb]
\centering
\includegraphics[width=\linewidth]{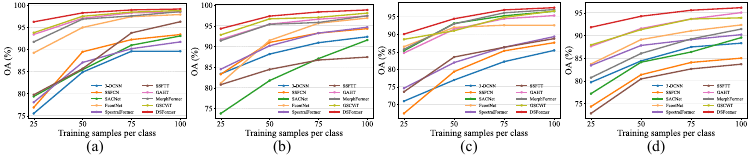}
\caption{Comparison of classification performance with different training sample per class: (a) Pavia University dataset. (b) Houston dataset. (c) Indian Pines dataset. (d) Whu-HongHu dataset.}
\label{fig:train_sample}
\end{figure*}

\textbf{(2) Performance Contribution of Spatial-Spectral Mechanism.}
We further explored the impact of the spatial-spectral selection mechanism within KSFTB on classification performance. As shown in Table \ref{table:ablation_2}, classification performance decreases in the four datasets when using the single spatial or spectral selection branch individually. This highlights the importance of considering both spatial and spectral information for the accurate identification of ground objects in HSIs. It is essential to not only adaptively optimal the spatial receptive field but also integrate the spectral features in appropriate scales, as the required spatial-spectral contextual information varies significantly across different land cover types. For example, in the Pavia University dataset, the OA reaches 96.09\% and 96.01\% respectively when using the single spatial or spectral branches. In contrast, when both spatial and spectral receptive field fusion mechanisms are applied simultaneously, the OA reaches 96.59\%. In summary, this ablation study demonstrates that optimal classification performance is achieved when both spatial and spectral multiscale features are employed, thus validating the effectiveness of the proposed KSFTB.

\subsection{Performance Comparison and Analysis}
In this subsection, we compare the proposed method with other classical and state-of-the-art approaches. A brief introduction to each method included in the experiments is provided below.

\begin{itemize}
  \item[1)] 3D-CNN \citep{3d-cnn}: Spatial-spectral classification using 3D-CNN. We follow the implementation with the original article.
  \item[2)] SSFCN \citep{ssfcn}: Spatial-spectral classification with an end-to-end fully convolutional network. The model is trained using the Adam optimizer with a learning rate of $5e-4$ and a momentum of 0.9.
  \item[3)] SACNet \citep{SACNet}: Spatial-spectral classification method that integrates self-attention learning with context encoding. The network is trained using the Adam optimizer with a learning rate of $5e-4$ and a weight decay of $5e-5$.
  \item[4)] FcontNet \citep{Fcnet}: Multiscale classification based on non-local contextual attention across space, bands, and scales. The base learning rate is 0.01, adjusted using the poly scheduling policy. The model is optimized using SGD with a momentum of 0.9 and a weight decay of $1e-4$.
  \item[5)] SpectralFormer \citep{SpectralFormer}: A Transformer-based architecture that employs sequential learning, focusing on adjacent bands and patches. The model is trained using the Adam optimizer with a learning rate of $5e-4$, which decays by a factor of 0.9 after every tenth of the total epochs.
  \item[6)] SSFTT \citep{SSFTT}: Spatial-spectral classification method integrates a CNN with a single-layer transformer and a Gaussian distribution-weighted tokenization module. The model is trained using the Adam optimizer with a learning rate of $1e-3$.
  \item[7)] GAHT \citep{GAHT}: Spatial-spectral classification through a hierarchical transformer with grouped pixel embedding. The model is trained using the SGD optimizer with a momentum of 0.9, a weight decay of $1e-4$, and a learning rate of $1e-3$.
  \item[8)] MorphFormer \citep{morphformer}: A ViT-based spatial-spectral classification method that combines attention mechanisms with morphological operations. The model is trained using the Adam optimizer with a weight decay of $5e-3$ and a learning rate of $5e-4$. 
  \item[9)] GSCViT \citep{GSCViT}: A lightweight network utilizing groupwise separable convolution and ViT in each feature extraction block. The model is trained with the AdamW optimizer, featuring a learning rate of $1e-3$ and a weight decay coefficient of 0.05.
\end{itemize}

\textbf{(1) Quantitative Results.} The quantitative results include the accuracy per class, OA, $\kappa$, and AA as shown in Table~\ref{table:PAVIA_OA} to Table~\ref{table:Indian_OA}, with the best results highlighted in bold. Taking the Pavia University dataset as an example, CNN-based HSI classification methods such as 3D-CNN at the patch level, the classification performance is limited by the size of the input patch, resulting in an OA of only 77.01\%. For image-based methods such as SSFCN, although the entire image is used as input, the limited receptive field of the convolutional layer hampers the effective extraction of contextual information, leading to a lower OA of 81.48\%. The SACNet method introduces a self-attention learning mechanism and context encoding to establish long-range dependencies, but its performance improvement is restricted by spatial resolution during the self-attention mechanism calculation, yielding an OA of 83.27\%. In contrast, FcontNet achieves a relatively higher accuracy with an OA of 90.54\% by combining scale, spatial, and channel attention mechanisms for multiscale classification. However, the extensive self-attention computations and multiscale feature design of this method result in high computational resource demands and low operational efficiency.

For transformer-based methods, both SpectralFormer and SSFTT are built on the ViT structure and learn by constructing sequence data from adjacent bands. However, due to redundancy between bands and the lack of spatial context modeling, these methods perform slightly worse than the CNN-based HSI classification methods aforementioned, with an OA of 76. 46\% and 75.79\%, respectively. GAHT builds a hierarchical transformer on the ViT foundation by integrating grouped pixel embeddings, MorphFormer incorporates morphological operations into the ViT attention mechanism, and GSCViT uses groupwise separable convolution with ViT to extract local and global features. All of these methods achieve relatively high accuracy, with OAs of 90.69\%, 90.6\%, and 93.40\%, respectively. However, these approaches overlook the varying contextual ranges required by different types of land cover in the HSI, and the traditional full self-attention mechanism will inevitably includes redundant and irrelevant information during computation. Furthermore, insufficient refinement in capturing the interspectral context limits the accuracy of existing transformer-based HSI classification methods.

To address the aforementioned issues, we employ KSFTB for adaptive receptive field selection and fusion, followed by TSFTB to intelligently choose the most relevant spatial-spectral tokens for self-attention computation. Our method, DSFormer, achieved an optimal OA of 96.59\%, surpassing the second-best method by 3.2\%. Notably, DSFormer obtained the best results in nearly all per-class accuracy metrics, except for the ninth class, shadows, where it was slightly outperformed by MorphFormer. This may be attributed to the integration of morphological operations within the attention mechanism, which likely provides more clear category priors for shadow recognition. Similarly, our proposed method also demonstrated superior accuracy on other additional datasets, highlighting its advantages and applicability.

\begin{figure}[t]
\centering
\includegraphics[width=\linewidth]{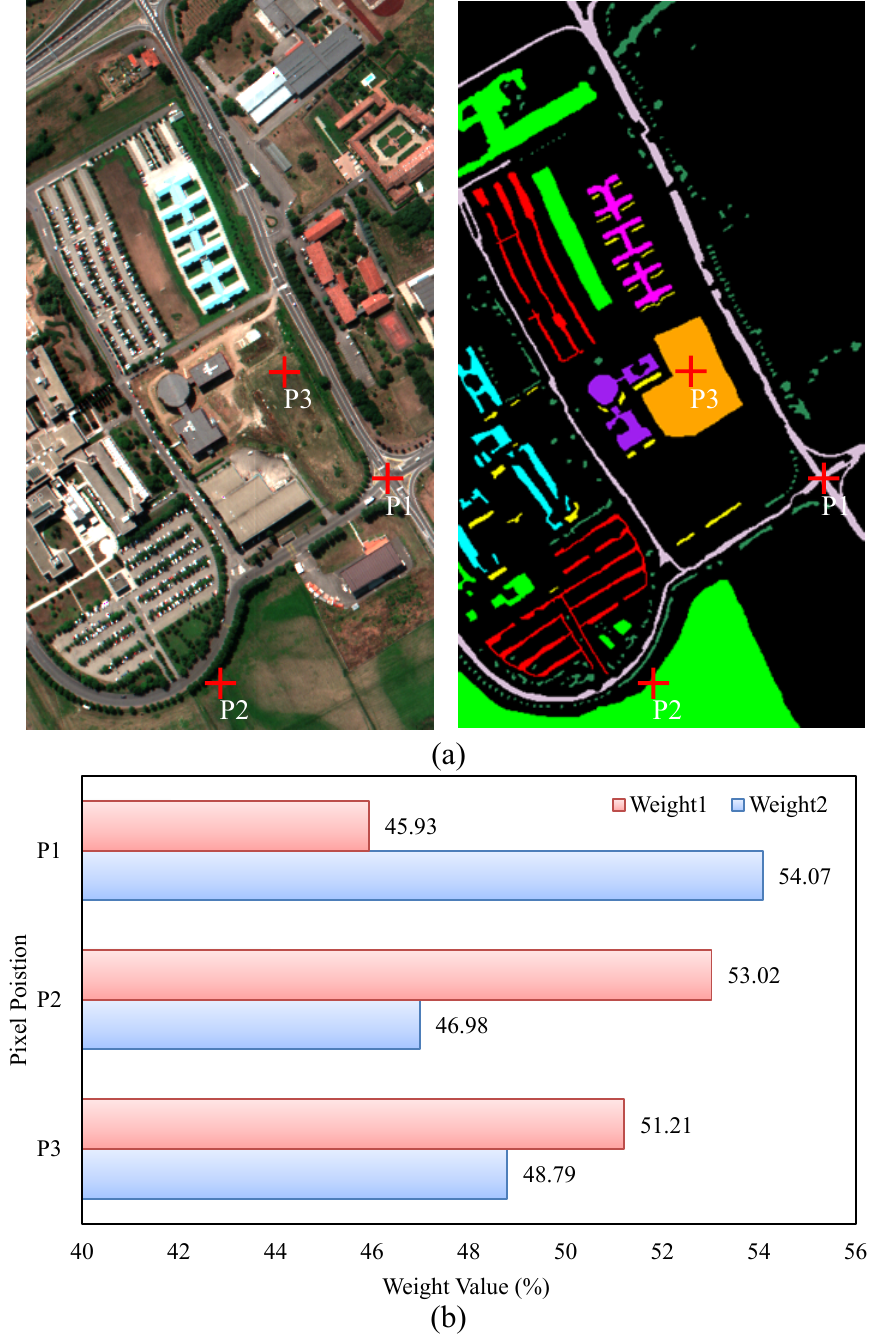}
\caption{Visualization of the objects with different receptive fields selection weights on the Pavia University dataset. (a) Position of three objects P1: Asphalt, P2: Meadows, and P3:Bare soil. (b) Weights selection values of different receptive fields.}
\label{fig:plot_w}
\end{figure}

\textbf{(2) Qualitative Results.} We also present classification maps in Fig.~\ref{fig:PaviaMap}-\ref{fig:IndianMap} to evaluate various methods. Using the Pavia University dataset as a case study, we observed that methods such as 3D-CNN and SSFCN exhibit significant salt-and-pepper noise. In contrast, FcontNet and the latest Transformer-based approaches yield smoother and better results. A particularly noteworthy observation is on the classification of the bricks class, where most existing methods exhibit omissions or misclassifications. However, our DSFormer model significantly reduces these errors, producing outputs that are much closer to the ground truth. This advantage has been consistently validated across multiple datasets, highlighting the superior performance of the DSFormer model in handling complex classification tasks.

\textbf{(3) Robustness with different training samples.} To assess the stability of the proposed DSFormer method, we conducted extensive experiments on four diverse datasets, each with varying numbers of training samples. Specifically, we randomly selected 25, 50, 75, and 100 samples per class to construct the training sets. Fig.~\ref{fig:train_sample} compares the classification performance in these different sample sizes. The results indicate that the classification accuracy of DSFormer is steadily improving as the number of training samples increases, demonstrating its robustness. Furthermore, DSFormer consistently outperforms existing state-of-the-art methods, achieving the highest accuracy across all training sample configurations.

\begin{table*}[!htb] 
\newcommand{\tabincell}[2]{\begin{tabular}{@{}#1@{}}#2\end{tabular}}
\centering
\caption{Comparison of model parameters (K), inference time (s) and OA (\%) across four datasets}\label{table:time}
\begin{tabular}{l|l|cccccc}

\toprule
Dataset                       & Method        & SpectralFormer & SSFTT      & GAHT       & MorphFormer & GSCViT      & DSFormer     \\

\midrule
\multirow{3}{*}{Pavia University}        & Parameters (K) & 174.19 & 484.42 & 927.11 & 149.75 & 984.33 & 571.88\\
                              & Inference time (s)  & 44.03 & 8.52 & 15.86 & 26.17 & 14.06 & 39.40\\
                              & OA (\%)       & 76.46 & 75.79 & 90.69 & 90.62 & 93.40 & 96.59\\ \hline
\multirow{3}{*}{Houston}      & Parameters (K)  & 235.95 & 673.74 & 973.58 & 155.19 & 995.22 & 577.90\\
                              & Inference time (s)   & 226.92 & 37.94 & 60.49 & 99.22 & 123.07 & 136.28\\
                              & OA (\%)       & 88.10 & 86.72 & 94.99 & 95.36 & 96.52 & 97.66\\ \hline
\multirow{3}{*}{Indian Pines} & Parameters (K)  & 352.44 & 931.85 & 830.80 & 199.99 & 1009.62 & 585.20 \\
                              & Inference time (s)   & 14.05 & 1.44 & 2.46 & 3.55 & 1.96 & 4.66\\
                              & OA (\%)       & 81.88 & 83.47 & 91.39 & 92.91 & 94.26 & 95.17\\ \hline
\multirow{3}{*}{Whu-HongHu}       & Parameters (K)  & 551.28 & 1254.80 & 1514.97 & 255.99 & 1027.93 & 594.93 \\
                              & Inference time (s)   & 274.36 & 45.98 & 63.54 & 83.65 & 49.25 & 108.51\\
                              & OA (\%)    & 86.36 & 77.17 & 91.65 & 91.19 & 91.79 & 94.59\\
\bottomrule
\end{tabular}
\end{table*}

\textbf{(4) Computation Cost Analysis.} Given that previous comparative experiments indicate that transformer-based models typically achieve higher accuracy, we performed a comprehensive comparison of these methods regarding parameters, inference time, and OA results, as presented in Table~\ref{table:time}. In our performance evaluation, the proposed DSFormer exhibited strong competitiveness and achieved a good balance between precision and efficiency in four benchmark datasets. Regarding model parameters, our DSFormer has fewer parameters than the state-of-the-art methods GAHT and GSCViT. 
For inference time, DSFormer is faster than SpectralFormer while achieving higher OAs. We believe the efficiency of DSFormer is primarily influenced by the grouping strategy in TSFTB, which generates a large number of tokens for subsequent self-attention calculations. In summary, these results suggest that DSFormer enables a good trade-off between computational resource usage and complex scene perception.

\begin{figure}[!htb]
\centering
\includegraphics[width=\linewidth]{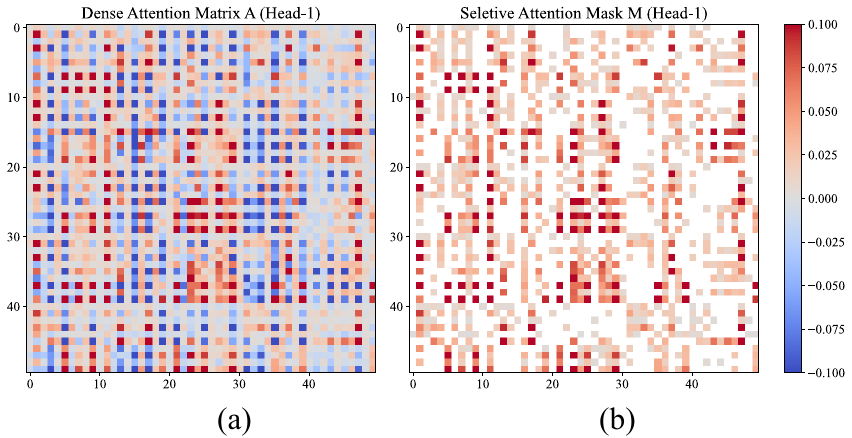}
\caption{Visualization of the alterations in the attention matrix before and after the token selective operation using the Pavia University dataset. (a) The dense attention matrix. (b) The selective attention mask. The attention map is obtained from the first head of MHSA with a group number of 2 and a selection rate of 40\%.}
\label{fig:plot_AM}
\end{figure}

\subsection{Visualization}

\textbf{(1) Different Objects of Selection Weights.}
To validate the differences in the receptive field requirements for various types of land cover in HSIs, we present visualizations of the selection weights for different objects during receptive field fusion. Fig.~\ref{fig:plot_w} (a) presents the locations of three land covers: Asphalt, Meadows and Bare Soil within the Pavia University dataset and their corresponding positions. And Fig.~\ref{fig:plot_w} (b) shows the weight values $\mathbf{W}_1$ and $\mathbf{W}_2$ in equation \ref{eqn:W} (normalized for visualization) for different receptive fields used in the final fusion within the KSFTB. The weight results show that the weight corresponding to the larger receptive field in P1 is 54.07\%, which confirms our previous analysis (see Fig. \ref{fig:Intro}) that asphalt requires more contextual information to perform an accurate classification. In contrast, at locations P2 and P3, both of which exhibit spatial continuity, smaller receptive field characteristics are sufficient for classification.

\textbf{(2) Attention Matrix Change of Token Selection.}
To further understand the token selection process in TSFTB, we present the changes in the attention matrix before and after the operation (corresponding to equation \ref{eqn:mask}). As shown in Fig.~\ref{fig:plot_AM}, we used the Pavia University dataset with a group number of 2 and a selection rate of 40\%. Specifically, we selected the dense attention matrix from the first head of the MHSA mechanism. From the figure, it is evident that after the token selection operation, only 40\% of the elements in the original dense attention matrix retain their values, while the remaining 60\% are set to 0, resulting in a sparse selective attention mask for subsequent effective fusion. Through this process, the relevant information for classification is retained, whereas irrelevant and redundant information is discarded.

\section{Conclusion}\label{Section5}
In this article, we introduce a novel network called DSFormer for HSI classification. DSFormer flexibly selects and fuses features across varying receptive fields, achieving contextual modeling of crucial spatial-spectral features through adaptive attention selection and fusion. Recognizing that objects in HSI of varying scales require distinct spatial-spectral contextual ranges, DSFormer incorporates a KSFTB, which enables the learning of optimal receptive fields by dynamically selecting and integrating multiscale features extracted through multiple dilated convolutions. Furthermore, since the original MHSA may introduce redundant and irrelevant regions and targets that negatively affect HSI classification performance. We develop a TSFTB that realize the adaptive selection of the most valuable spatial-spectral tokens during the context fusion process under a spectral grouping strategy. These selection and fusion mechanisms collectively enable DSFormer to achieve effective visual perception of spatial-spectral contexts within an optimal receptive field. Extensive experiments conducted on four benchmark HSI datasets demonstrate that the proposed DSFormer achieves competitive comprehension performance on accuracy, robustness and inference efficiency compared to existing advanced methods.


\bibliography{DSFormer}


\end{sloppypar}
\end{document}